\documentclass{article}

\usepackage[preprint]{nips_2018}

\usepackage[utf8]{inputenc} 
\usepackage[T1]{fontenc}    
\usepackage{hyperref}       
\usepackage{url}            
\usepackage{booktabs}       
\usepackage{amsfonts}       
\usepackage{nicefrac}       
\usepackage{microtype}      

\usepackage{amsmath}
\usepackage{amsthm}
\usepackage{mathtools}
\usepackage{algorithm,algorithmic}
\usepackage{subcaption}
\usepackage{tikz}
\usepackage{todo}
\usepackage{natbib}
\usetikzlibrary{arrows}

\DeclareMathOperator*{\median}{median}
\DeclareMathOperator*{\maximize}{maximize}
\DeclareMathOperator*{\minimize}{minimize}
\DeclareMathOperator*{\diag}{diag}
\DeclareMathOperator*{\subjectto}{subject\;to}
\DeclareMathOperator*{\for}{for}
\DeclareMathOperator*{\hardtanh}{hardtanh}

\newtheorem*{theorem*}{Theorem}
\newtheorem{theorem}{Theorem}
\newtheorem{lemma}{Lemma}
\newtheorem{corollary}{Corollary}

\newtheorem{definition}{Definition}

\title{Scaling provable adversarial defenses}

\author{
  Eric Wong\\
  Machine Learning Department\\
  Carnegie Mellon University\\
  Pittsburgh, PA 15213 \\
  \texttt{ericwong@cs.cmu.edu} \\
  \And
  Frank R. Schmidt\\
  Bosch Center for Artificial Intelligence \\
  Renningen, Germany \\
  \texttt{frank.r.schmidt@de.bosch.com} \\
  \And
  Jan Hendrik Metzen\\
  Bosch Center for Artificial Intelligence \\
  Renningen, Germany \\
  \texttt{janhendrik.metzen@de.bosch.com} \\
  \And
  J. Zico Kolter\\
  Computer Science Department \\
  Carnegie Mellon University and \\
  Bosch Center for Artificial Intelligence \\
  Pittsburgh, PA 15213 \\
  \texttt{zkolter@cs.cmu.edu} \\
}

\begin{document}

\maketitle

\begin{abstract}
Recent work has developed methods for learning deep network classifiers that are
\emph{provably} robust to norm-bounded adversarial perturbation; however, these
methods are currently only possible for relatively small feedforward networks. 
In this paper, in an effort to scale these approaches to substantially larger
models, we extend previous work in three main directions.  First, we present a
technique for extending these training procedures to much more general networks,
with skip connections (such as ResNets) and general nonlinearities; the
approach is fully modular, and can be implemented automatically (analogous to
automatic differentiation). Second, in the specific case of $\ell_\infty$ adversarial
perturbations and networks with ReLU nonlinearities, we adopt a nonlinear random
projection for training, which scales \emph{linearly} in the number of
hidden units (previous approaches scaled quadratically).  Third, we show how to
further improve robust error through cascade models.  On both MNIST and CIFAR
data sets, we train classifiers that improve substantially on the state of the
art in provable robust adversarial error bounds: from 5.8\% to 3.1\% on MNIST 
(with $\ell_\infty$ perturbations of $\epsilon=0.1$), and from 80\% to 36.4\% on
CIFAR (with $\ell_\infty$ perturbations of $\epsilon=2/255$). Code for all experiments 
in the paper is available at \url{https://github.com/locuslab/convex_adversarial/}. 

\end{abstract}

\section{Introduction}

A body of recent work in adversarial machine learning has shown that it is
possible to learn \emph{provably robust} deep classifiers 
\citep{wong2017provable,raghunathan2018certified,dvijotham2018dual}.  These are
deep networks that are verifiably \emph{guaranteed} to be robust to adversarial
perturbations under some specified attack model; for example, a certain
robustness certificate may guarantee that for a given example $x$, no
perturbation $\Delta$ with $\ell_\infty$ norm less than some specified
$\epsilon$ could change the class label that the network predicts for the
perturbed example $x + \Delta$.  However, up until this point, such provable
guarantees have only been possible for reasonably small-sized networks.  It has remained unclear 
whether these methods could extend to larger, more representionally
complex networks.

In this paper, we make substantial progress towards the goal of scaling these
provably robust networks to realistic sizes.  Specifically, we
extend the techniques of \citet{wong2017provable} in three key ways.  First,
while past work has only applied to pure feedforward networks, we extend the
framework to deal with arbitrary residual/skip connections (a hallmark of modern
deep network architectures), and arbitrary activation functions 
(\citet{dvijotham2018dual} also worked with arbitrary activation functions, but
only for feedforward networks, and just discusses network verification rather
than robust training).  Second, and possibly most importantly, computing the upper  
bound on the robust loss in \citep{wong2017provable} in the worst case scales 
\emph{quadratically} in the number of hidden units in the network, making the
approach impractical for larger networks.  In this work, we use a nonlinear
random projection technique to estimate the bound in manner that scales only
linearly in the size of the hidden units (i.e., only a constant multiple times
the cost of traditional training), and which empirically can be used to train the 
networks with no degradation in performance from the previous work. 
Third, we show how to further improve robust performance of these methods,
though at the expense of worse non-robust error, using multi-stage cascade
models. Through these extensions, we are able to improve substantially upon the verified robust errors obtained 
by past work.

\section{Background and related work}
Work in adversarial defenses typically falls in one of three primary categories.
First, there is ongoing work in developing heuristic defenses against
adversarial examples: 
\citep{goodfellow2015explaining,papernot2016distillation,kurakin2016adversarial,metzen2017detecting}
to name a few.  While this work is largely empirical at this
point, substantial progress has been made towards developing networks that seem
much more robust than previous approaches.  Although a distressingly large
number of these defenses are quickly ``broken'' by more advanced attacks 
\citep{athalye2018obfuscated}, there have also been some methods that have proven empirically resistant 
to the current suite of attacks; the recent NIPS 2017 adversarial
example challenge \citep{kurakin2018adversarial}, for example, highlights some of the progress made on
developing classifiers that appear much stronger in practice than many of the
ad-hoc techniques developed in previous years.  Many of the approaches,
though not formally verified in the strict sense during training, nonetheless
have substantial theoretical justification for why they may perform well: 
\citet{sinha2018certifiable} uses properties of statistical robustness to
develop an approach that is not much more difficult to train and which
empirically does achieve some measure of resistance to attacks; \citet{madry2017towards}
considers robustness to a first-order adversary, and shows that a randomized
projected gradient descent procedure is optimal in this setting.
Indeed, in some cases the classifiers trained via these methods can be
verified to be adversarially robust using the verification techniques
discussed below (though only for very small
networks). Despite this progress, we believe it is also crucially important to
consider defenses that \emph{are} provably robust, to avoid any possible attack.

Second, our work in this paper relates closely to techniques for the formal
verification of neural networks systems (indeed, our approach can be viewed as
a convex procedure for verification, coupled with a method for training
networks via the verified bounds).  In this area, most past work focuses on
using exact (combinatorial) solvers to verify the robustness properties of
networks, either via Satisfiability Modulo Theories (SMT) solvers 
\citep{huang2017safety, ehlers2017formal, carlini2017towards} or integer
programming approaches \citep{lomuscio2017approach,
tjeng2017verifying,cheng2017maximum}.  These methods have the benefit of being able to reason
exactly about robustness, but at the cost of being combinatorial in complexity.
This drawback has so far prevented these methods from effectively scaling to
large models or being used within a training setting.  There have also been a
number of recent attempts to verify networks using non-combinatorial methods 
(and this current work fits broadly in this general area).  For example, 
\citet{gehr2018ai} develop a suite of verification methods based upon abstract
interpretations (these can be broadly construed as relaxations of combinations
of activations that are maintained as they pass through the network). 
\citet{dvijotham2018dual} use an approach based upon analytically solving an
optimization problem resulting from dual functions of the activations (which
extends to activations beyond the ReLU). However, these
methods apply to simple feedforward architectures without skip connections, 
and focus only on verification of existing networks. 

Third, and most relevant to our current work, there are several approaches that
go beyond provable verification, and also integrate the verification procedure
into the training of the network itself.  For example, \citet{hein2017formal}
develop a formal bound for robustness to $\ell_2$ perturbations in two-layer
networks, and train a surrogate of their bounds.  
\citet{raghunathan2018certified} develop a semidefinite programming (SDP)
relaxation of exact verification methods, and train a network by minimizing this
bound via the dual SDP.  And \citet{wong2017provable} present a
linear-programming (LP) based upper bound on the robust error or loss that can
be suffered under norm-bounded perturbation, then minimize this upper bound
during training; the method is particularly efficient since they do not solve
the LP directly, but instead show that it is possible to bound the LP optimal
value and compute elementwise bounds on the activation functions based on a
backward pass through the network.  However, it is still the case that none of
these approaches scale to realistically-sized networks; even the approach of 
\citep{wong2017provable}, which empirically has been scaled to the largest
settings of all the above approaches, in the worst case scales 
\emph{quadratically} in the number of hidden units in the network and dimensions
in the input.  Thus, all the approaches so far have been limited to relatively
small networks and problems such as MNIST.

\paragraph{Contributions}  This paper fits into this third
category of integrating verification into training, and makes substantial progress towards scaling these methods to
realistic settings.  While we cannot yet reach e.g. ImageNet scales, even in
this current work, we show that it \emph{is} possible to overcome the main
hurdles to scalability of past approaches.  Specifically, we develop a provably
robust training procedure, based upon the approach in \citep{wong2017provable},
but extending it in three key ways.  The resulting method: 1) extends to general
networks with skip connections, residual layers, and activations besides the
ReLU; we do so by using a general formulation based on the Fenchel conjugate
function of activations; 2) scales \emph{linearly} in the dimensionality of the
input and number of hidden units in the network, using techniques from nonlinear
random projections, all while suffering minimal degradation in accuracy;  and 3)
further improves the quality of the bound with model cascades.  We describe each
of these contributions in the next section.

\section{Scaling provably robust networks}

\subsection{Robust bounds for general networks via modular dual functions}
This section presents an architecture for constructing provably robust bounds
for general deep network architectures, using Fenchel duality. Importantly,
we derive the dual of each network operation in a fully modular fashion, 
simplifying the
problem of deriving robust bounds of a network to 
bounding the dual of individual functions. By building up a toolkit of dual 
operations, we can automatically construct the dual of any network architecture 
by iterating through the layers of the original network.  

\subparagraph{The adversarial problem for general networks}
We consider a generalized $k$ ``layer'' neural network $f_\theta : \mathbb{R}^{|x|}\rightarrow \mathbb{R}^{|y|}$ given by the equations 
\begin{equation}
\label{eq:nn_form}
  \begin{split}
    z_{i} & = \sum_{j=1}^{i-1} f_{ij}(z_j), \;\;\for i=2,\ldots,k\\
  \end{split}
\end{equation}
where $z_1 = x$, $f_\theta(x) \equiv z_k$ (i.e., the output of the network) and
$f_{ij} : \mathbb{R}^{|z_j|} \rightarrow \mathbb{R}^{|z_i|}$ is some function from layer $j$ to layer $i$.
Importantly, this differs from prior work in two key ways. First, unlike the conjugate forms 
found in \citet{wong2017provable, dvijotham2018dual}, we no longer assume that the network 
consists of linear operations followed by activation functions, and instead opt to work with an 
arbitrary sequence of $k$ functions. This simplifies the analysis of sequential non-linear 
activations commonly found in modern architectures, e.g. max pooling or a
normalization strategy followed by a ReLU,\footnote{Batch normalization\cite{},
since it depends on entire minibatches, is formally not covered by the
approach, but it can be approximated by considering the scaling and shifting to
be generic parameters, as is done at test time.} by analyzing each activation
independently, whereas previous work would need to analyze the entire sequence
as a single, joint activation.  Second, we allow layers to depend not just on
the previous layer, but also on all layers before it. This generalization
applies to networks with any kind of skip connections, e.g. residual networks
and dense networks, and greatly expands the set of possible architectures. 

Let $\mathcal B(x) \subset \mathbb{R}^{|x|}$, represent some input constraint for the adversary. 
For this section we will focus on an arbitrary norm ball 
$\mathcal{B}(x) = \{ x + \Delta : \|\Delta\| \leq \epsilon\}$. This is the constraint set 
considered for norm-bounded adversarial perturbations, however other constraint sets can 
certainly 
be considered. Then, given an input example $x$, a known label $y^*$, and a
target label $y^{\mathrm{targ}}$, the problem of finding the most adversarial
example within $\mathcal{B}$ (i.e., a so-called \emph{targeted} adversarial
attack) can be written as
\begin{equation}
  \label{eq:primal}
    \minimize_{z_k} \;\; c^T z_k, \;\; 
    \subjectto \;\; 
    z_{i}  = \sum_{j=1}^{i-1} f_{ij}(z_j), \;\;\for i=2,\ldots,k, \;\;
    z_1 \in \mathcal{B}(x)
\end{equation}
where $c=e_{y^\star} - e_{y^{\mathrm{targ}}}$. 
\paragraph{Dual networks via compositions of modular dual functions}
To bound the adversarial problem, we look to its dual optimization problem using the machinery 
of Fenchel conjugate functions \citep{fenchel1949conjugate}, described in Definition \ref{def:conjugate}. 
\begin{definition}
\label{def:conjugate}
The conjugate of a function $f$ is another function $f^*$ defined by 
\begin{equation}
f^*(y) = \max_x x^Ty - f(x)
\end{equation}
\end{definition}
Specifically, we can lift the constraint $z_{i+1} = \sum_{j=1}^i f_{ij}(z_j)$
from Equation \ref{eq:primal} into the objective with an indicator 
function, and use conjugate functions to obtain a lower bound. 
For brevity, we will use the subscript notation $(\cdot)_{1:i} = ((\cdot)_1, \dots,(\cdot)_i)$, e.g. $z_{1:i} = (z_1, \dots, z_i)$. 
Due to the skip connections, the indicator functions are not independent, so 
we cannot directly conjugate each individual indicator function. We can, however, 
still form its 
dual using the conjugate of a different indicator 
function corresponding to the backwards direction, as shown in Lemma \ref{cor:support}. 
\begin{lemma}
\label{cor:support}
Let the indicator function for the $i$th constraint be 
\begin{equation}
\chi_{i}(z_{1:i}) = \left\{\begin{array}{ll}
0 & \text{if }z_{i} = \sum_{j=1}^{i-1} f_{ij}(z_j)\\
\infty & \text{otherwise,}
\end{array} \right.
\end{equation}
for $i=2, \dots, k$, and consider the joint indicator function $\sum_{i=2}^k \chi_i(z_{1:i})$. Then, the joint indicator is lower bounded by $\max_{\nu_{1:k}} \nu_k^Tz_k -\nu_1^Tz_1 - \sum_{i=1}^{k-1} \chi_{i}^*(-\nu_i, \nu_{i+1:k})$, where  
\begin{equation}
\chi_{i}^*(\nu_{i:k}) = \max_{z_i} \nu_i^Tz_i + \sum_{j=i+1}^{k}\nu_{j}^Tf_{ji}(z_i)
\end{equation}
for $i=1, \dots, k-1$. Note that $\chi_{i}^*(\nu_{i:k})$ is the exact conjugate of the  indicator for the set $\{x_{i:k} : x_{j} = f_{ji}(x_i)\;\;\forall j > i\}$, which is 
different from the set indicated by $\chi_i$. However, when \\
there are no skip connections (i.e. $z_i$ only depends on $z_{i-1}$), 
$\chi_i^*$ is exactly the conjugate of $\chi_i$. 
\end{lemma}
We defer the proof of Lemma \ref{cor:support} to Appendix \ref{app:conjugates}. With structured upper bounds 
on these conjugate functions, we can bound the original adversarial problem 
using the dual network described in Theorem \ref{thm:dual_network}. 
We can then optimize the bound using any standard deep learning toolkit 
using the same robust optimization procedure as in \citet{wong2017provable} 
but using our bound instead. 
This amounts to minimizing the loss evaluated on our bound of possible network outputs 
under perturbations, as a drop in replacement for the traditional network output. 
For the adversarial setting, note 
that the $\ell_\infty$ perturbation results in a dual norm of $\ell_1$. 

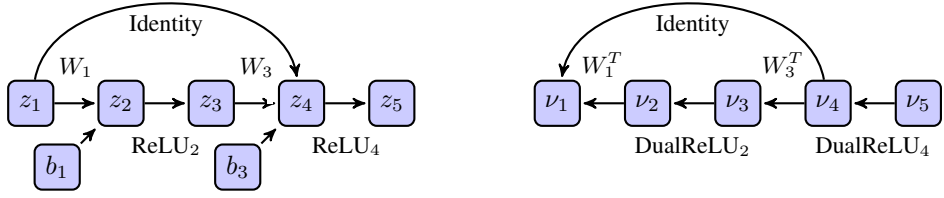
\begin{figure}
  \centering
  \begin{subfigure}{0.9\columnwidth}
  \begin{tikzpicture}[->,>=stealth',shorten >=1pt,node distance=1.2cm,
  thick,main node/.style={rounded corners=0.1cm,fill=blue!20,draw,
  minimum size=6mm}]

  \node[main node] (z5) {$z_5$};
  \node[main node] (z4) [left of=z5] {$z_4$};
  \node[main node] (z3) [left of=z4] {$z_3$};
  \node[main node] (b3) [below left of=z4] {$b_3$};
  \node[main node] (z2) [left of=z3] {$z_2$};
  \node[main node] (z1) [left of=z2] {$z_1$};
  \node[main node] (b1) [below left of=z2] {$b_1$};

  \path[every node/.style={font=\small,
  		fill=white,inner sep=1pt}]
    (b1) edge [] node[above=4mm] {} (z2)
    (z1) edge [] node[above=3mm] {$W_1$} (z2)
         edge [bend left=80] node[below=1mm] {$\text{Identity}$} (z4)
    (z2) edge [] node[below=4mm] {$\text{ReLU}_2$} (z3)
    (z3) edge [] node[above=3mm] {$W_3$} (z4)
    (b3) edge [] node[above=4mm] {} (z4)
    (z4) edge [] node[below=4mm] {$\text{ReLU}_4$} (z5);

  \node[main node] (nu1) [right of=z5, xshift=1cm] {$\nu_1$};
  \node[main node] (nu2) [right of=nu1] {$\nu_2$};
  \node[main node] (nu3) [right of=nu2] {$\nu_3$};
  \node[main node] (nu4) [right of=nu3] {$\nu_4$};
  \node[main node] (nu5) [right of=nu4] {$\nu_5$};

  \path[every node/.style={font=\small,
  		fill=white,inner sep=1pt}]
    (nu5) edge [] node[below=4mm] {$\text{DualReLU}_4$} (nu4)
    (nu4) edge [] node[above=3mm] {$W_3^T$} (nu3)
          edge [bend right=80] node[below=1mm] {$\text{Identity}$} (nu1)
    (nu3) edge [] node[below=4mm] {$\text{DualReLU}_2$} (nu2)
    (nu2) edge [] node[above=3mm] {$W_1^T$} (nu1);
\end{tikzpicture}
\end{subfigure}
  \caption{An example of the layers forming a typical residual block (left) and its dual (right), 
  using the dual layers described in Corollaries \ref{cor:linear} and \ref{cor:relu}. 
	Note that the bias terms of the residual network go into the dual objective and
	are not part of the structure of the dual network, and the skip connections
	remain in the dual network but go in the opposite direction. }
\label{fig:example_dual}
\end{figure}

\begin{theorem}
\label{thm:dual_network}
Let $g_{ij}$ and $h_i$ be any functions such that 
\begin{equation}
\chi_{i}^*(-\nu_i, \nu_{i+1:k}) \leq h_i(\nu_{i:k}) \;\; \subjectto \;\; \nu_i = \sum_{j=i+1}^{k}g_{ij}(\nu_{j})
\label{eq:conj_bound}
\end{equation}
for $i=1, \dots, k-1$. Then, the adversarial problem from Equation \ref{eq:primal} is lower bounded by 
\begin{equation}
\label{eq:dual_obj}
J(x, \nu_{1:k}) = -\nu_1^Tx - \epsilon\|\nu_1\|_* - \sum_{i=1}^{k-1} h_i(\nu_{i:k})
\end{equation}
where $\|\cdot\|_*$ is the dual norm, and $\nu_{1:k}=g(c)$ is the output of a $k$ layer neural network $g$ on input $c$, given by the equations 
\begin{equation}
\label{eq:dual_network}
\nu_k = -c, \;\; \nu_i = \sum_{j=i}^{k-1}g_{ij}(\nu_{j+1}), \;\; \for i=1,\dots,k-1.
\end{equation}
\end{theorem}
We denote the upper bound on the conjugate function from Equation
\ref{eq:conj_bound} 
a \emph{dual layer}, and defer the proof to Appendix \ref{app:thm1}. To give a concrete example, we present 
two possible  
dual layers for linear operators and ReLU activations in 
Corollaries \ref{cor:linear} and \ref{cor:relu} (their derivations
are in Appendix \ref{app:dual_layers}), and we also depict 
an example dual residual block in Figure \ref{fig:example_dual}. 

\begin{corollary}
\label{cor:linear}
The dual layer for a linear operator $\hat z_{i+1} = W_iz_i + b_i$ is 
\begin{equation}
\chi_{i}^*(\nu_{i:k}) = \nu_{i+1}^Tb_i \;\; \subjectto \;\; \nu_i = W_i^T\nu_{i+1}.
\end{equation}
\end{corollary}
\begin{corollary}
\label{cor:relu}
Suppose we have lower and upper bounds $\ell_{ij}, u_{ij}$ on 
the pre-activations. 
The dual layer for a ReLU activation $\hat z_{i+1} = \max(z_i,0)$ is 
\begin{equation}
\chi_{i}^*(\nu_{i:k}) \leq - \sum_{j\in \mathcal I_i} \ell_{i,j} [\nu_{ij}]_+ \;\; \subjectto \;\; \nu_i = D_i\nu_{i+1}.
\end{equation}
where $\mathcal I^-_i, \mathcal I^+_i, \mathcal I$ denote the index sets where the bounds are negative, positive or spanning the origin respectively, and where $D_i$ is a diagonal matrix with entries 
\begin{equation}
\label{eq:d_after_alpha}
  \begin{split}
	(D_i)_{jj} & = \left \{
      \begin{array}{ll}
        0 & j \in \mathcal{I}^{-}_i \\
        1 & j \in \mathcal{I}^{+}_i \\
        \frac{u_{i,j}}{u_{i,j} - \ell_{i,j}} & j \in \mathcal{I}_i \\
      \end{array} \right .
   \end{split} .
 \end{equation}
\end{corollary}

We briefly note that these dual layers recover the original 
dual network described in \citet{wong2017provable}. 
Furthermore, the dual linear operation is 
the exact conjugate and introduces no looseness to the bound, 
while the dual ReLU uses the same relaxation used in \citet{ehlers2017formal, wong2017provable}. More generally, the strength of the bound from Theorem 
\ref{thm:dual_network} relies entirely on the tightness 
of the individual dual layers to their respective conjugate functions in Equation \ref{eq:conj_bound}. 
While any $g_{ij}$, $h_i$ can be chosen to upper bound the conjugate function, 
a tighter bound on the conjugate results in a tighter bound on the adversarial
problem. 

If the dual layers for all operations are linear, 
the bounds for all layers can be computed with a single forward pass 
through the dual network using a direct generalization of the form used in \citet{wong2017provable} (due to their similarity, 
we defer the exact algorithm to Appendix \ref{app:autodual}). 
By trading off tightness of the bound 
with computational efficiency by using 
linear dual layers, we can efficiently compute 
all bounds and construct the dual network one layer at a time. 
The end result is that 
we can automatically construct dual networks from dual layers in a fully modular 
fashion, completely independent of the overall 
network architecture 
(similar to how auto-differentiation tools proceed one 
function at a time to compute all parameter gradients using only 
the local gradient of each function). 
With a sufficiently comprehensive toolkit of dual layers, we can compute 
provable bounds on the adversarial problem for any network architecture. 

For other dual layers, we point the reader to two resources. For the explicit 
form of dual layers 
for hardtanh, batch normalization, residual connections, 
we direct the reader to Appendix \ref{app:dual_layers}. 
For analytical forms of conjugate functions of other activation 
functions such as tanh, sigmoid, and max pooling, we refer the 
reader to \citet{dvijotham2018dual}. 


\subsection{Efficient bound computation for $\ell_\infty$ perturbations via random projections}
A limiting factor of the proposed algorithm and the work of \citet{wong2017provable} is its 
computational complexity: for instance, to compute the bounds exactly for
$\ell_\infty$ norm bounded perturbations in ReLU networks, it is computationally
expensive to calculate $\|\nu_1\|_1$ and 
$\sum_{j\in \mathcal I_i}\ell_{ij}[\nu_{ij}]_+$. In contrast to other terms like $\nu_{i+1}^Tb_i$ 
which require only sending a single bias vector through the dual network, the matrices $\nu_1$ 
and $\nu_{i,\mathcal I_i}$ must be explicitly formed by sending an example through the dual network 
for each input dimension and for each $j \in \mathcal I_i$, which renders the
entire computation \emph{quadratic} in the number of hidden units. To scale the
method for larger, ReLU networks with $\ell_\infty$ perturbations, 
we look to random Cauchy 
projections.  Note that for an $\ell_2$ norm bounded adversarial perturbation,
the dual norm is also an $\ell_2$ norm, so we can use traditional random
projections \citep{vempala2005random}.  Experiments for the $\ell_2$ norm are 
explored further in Appendix \ref{app:l2}.  
However, for the remainder of this
section we focus on the $\ell_1$ case arising from $\ell_\infty$ perturbations.

\paragraph{Estimating with Cauchy random projections}
From the work of \citet{li2007nonlinear}, we can use the sample median estimator with Cauchy random projections to directly estimate $\|\nu_1\|_1$ for 
linear dual networks, and use 
a variation to estimate $\sum_{j\in \mathcal I}\ell_{ij}[\nu_{ij}]_+$, as shown in Theorem \ref{thm:projection} (the proof is in Appendix \ref{app:thm2}). 

\begin{theorem}
\label{thm:projection}. 
Let $\nu_{1:k}$ be the dual network from Equation \ref{thm:dual_network} 
with linear dual layers and let $r >0$ be the projection dimension. Then, we can estimate
\begin{equation}
\|\nu_1\|_1 \approx \median(|\nu_1^TR|)
\end{equation}
where $R$ is a $|z_1| \times r$ standard Cauchy random matrix and the median is taken over the 
second axis. Furthermore, we can estimate
\begin{equation}
\label{eq:d}
\sum_{j\in \mathcal I}\ell_{ij}[\nu_{ij}]_+ \approx \frac{1}{2}\left(-\median
(|\nu_i^T\diag(d_i)R|) + \nu_i^Td_i\right), \;\; 
d_{i,j} = \left\{\begin{array}{ll}
\frac{u_{i,j}}{u_{i,j} - \ell_{i,j}} & j \not \in \mathcal I_i\\
0 & j \in \mathcal I_i \end{array} \right .
\end{equation}
where $R$ is a $|z_i| \times r$ standard Cauchy random matrix, and the median is
taken over the second axis.
\end{theorem}

This estimate has two main advantages: first, it is simple to compute, as evaluating $\nu_1^TR$ 
involves passing the random matrix forward 
through the dual network (similarly, the other term requires passing a modified random matrix through the dual network; the exact algorithm is detailed in \ref{alg:estimate}). Second, it 
is memory efficient in the backward pass, as the gradient need only propagate through the median entries. 

\begin{algorithm}[tb]
   \caption{Estimating $\|\nu_1\|_1$ and $\sum_{j\in \mathcal I}\ell_{ij}[\nu_{ij}]_+$}
   \label{alg:estimate}
\begin{algorithmic}
  \STATE \textbf{input:} Linear dual network operations $g_{ij}$, projection dimension $r$, lower bounds $\ell_{ij}$, $d_{ij}$ from Equation \ref{eq:d}, layer-wise sizes $|z_i|$
  \STATE $R_1^{(1)} := \text{Cauchy}(r,|z_1|)$\; \emph{// initialize random matrix for $\ell_1$ term}
  \FOR{$i=2,\ldots,k$}
  \STATE \emph{// pass each term forward through the network}
  \FOR{$j=1,\dots,i-1$}
  \STATE $R_j^{(i)}, S_j^{(i)} := \sum_{k=1}^{i-1} g_{ki}^T(R_i^{(k)}), \sum_{k=1}^{i-1} g_{ki}^T(S_i^{(k)})$
  \ENDFOR
  \STATE $R_i^{(i)}, S_i^{(i)} := \diag(d_i)\text{Cauchy}(|z_i|,r), d_i$ \emph{// initialize terms for layer $i$}
  \ENDFOR
  \STATE \textbf{output:} $\median(|R_1^{(k)}|), 0.5\left(-\median(|R_2^{(k)}|) + S_2^{(k)}\right), \dots, 0.5\left(-\median(|R_{k}^{(k)}|) + S_{k}^{(k)}\right)$
\end{algorithmic}
\end{algorithm}

These random projections reduce the computational complexity of computing these terms to piping 
$r$ random Cauchy vectors (and an additional vector) through the network. Crucially, the complexity is no longer a quadratic function of the 
network size: if we fix the projection dimension to some constant $r$, then the computational 
complexity is now linear with the input dimension and $\mathcal I_i$. Since previous work was 
either quadratic or combinatorially expensive to compute, estimating the bound with random 
projections is the fastest and most scalable approach towards training robust networks that we 
are aware of. At test time, the bound can be computed exactly, as the gradients 
no longer need to be stored. However, if desired, it is possible to use a 
different estimator (specifically, the geometric estimator) for the $\ell_\infty$ norm to calculate high probability 
bounds on the adversarial problem, which is discussed in Appendix \ref{app:high_prob}.

\subsection{Bias reduction with cascading ensembles}

A final major challenge of training models to minimize a robust bound on the
adversarial loss, is that the robustness penalty acts as a regularization.  For
example, in a two-layer ReLU network, the robust loss penalizes $\epsilon \|\nu_1\|_1
= \epsilon \|W_1 D_1 W_2\|_1$, which effectively acts as a regularizer on the
network with weight $\epsilon$.  Because of this, the resulting networks (even
those with large representational capacity), are typically overregularized to
the point that many filters/weights become identically zero (i.e., the network
capacity is not used).

To address this point, we advocate for using a robust \emph{cascade} of
networks: that is, we train a sequence of robust classifiers, where later
elements of the cascade are trained (and evaluated) \emph{only on those
examples that the previous elements of the cascade cannot certify} (i.e., those
examples that lie within $\epsilon$ of the decision boundary). This
procedure is formally described in the Appendix in Algorithm 
\ref{alg:cascade}.

\section{Experiments}

\subparagraph{Dataset and Architectures} We evaluate the techniques in this
paper on two main datasets: MNIST digit classification \citep{lecun1998gradient}
and CIFAR10 image
classification \citep{krizhevsky2009learning}.\footnote{We fully realize the
irony of a paper with ``scaling" in the title that currently maxes out on
CIFAR10 experiments. But we emphasize that when it comes to certifiably robust
networks, the networks we consider here, as we illustrate below in Table 
\ref{table:sizes}, 
are more than an order of magnitude larger than any that have been considered
previously in the literature. Thus, our emphasis is really on the potential
scaling properties of these approaches rather than large-scale experiments on
e.g. ImageNet sized data sets.}
We test on a variety of deep and wide convolutional architectures, with and without residual connections.  All code for these experiments is
available at  \url{https://github.com/locuslab/convex_adversarial/}. 
The small network is the same as that used in \citep{wong2017provable}, with two convolutional layers of 16 and 32 filters and a fully connected layer of 100 units. The large network is a scaled up version of it, with four convolutional layers with 32, 32, 64, and 64 filters, and two fully connected layers of 512 units. The residual networks use the same structure used by \citep{zagoruyko2016wide} with 4 residual blocks with 16, 16, 32, and 64 filters. 
We highlight a subset of the results in Table \ref{table:results}, and briefly  
describe a few key observations below. We leave more extensive 
experiments and details regarding
the experimental setup in Appendix \ref{app:results}, including
additional experiments on $\ell_2$ perturbations. All  results
except where otherwise noted use random projection of 50 dimensions.

\begin{table}
\centering
\caption{Number of hidden units, parameters, and time per epoch for various architectures.}
\label{table:sizes}
\tabcolsep=0.11cm
\begin{tabular}{llrrr}
\textbf{Model} & \textbf{Dataset} & \textbf{\# hidden units} & \textbf{\# parameters} & \textbf{Time (s) / epoch} \\
\hline
Small  & MNIST & 4804 & 166406 & 74 \\
       & CIFAR & 6244 & 214918 & 48  \\
\hline
Large  & MNIST & 28064 & 1974762 & 667 \\
       & CIFAR & 62464 & 2466858 & 466 \\
\hline
Resnet & MNIST & 82536 & 3254562 & 2174 \\
       & CIFAR & 107496 & 4214850 & 1685 \\
\hline
\end{tabular}
\end{table}

\begin{table}
\centering
\caption{Results on MNIST, and CIFAR10 with small networks, large networks,
residual networks, and cascaded variants.}
\label{table:results}
\tabcolsep=0.11cm
\begin{tabular}{lll|rr|rr}
 &  &  & \multicolumn{2}{c|}{\textbf{Single model error}} & \multicolumn{2}{c}{\textbf{Cascade error}} \\
\textbf{Dataset} & \textbf{Model} & \textbf{Epsilon} & \textbf{Robust} & \textbf{Standard} &  \textbf{Robust} & \textbf{Standard}\\
\hline
MNIST & Small, Exact & 0.1 & 4.48\% & 1.26\% & -& - \\
MNIST & Small & 0.1 & 4.99\% & 1.37\% & \textbf{3.13\%} & \textbf{3.13}\% \\
MNIST & Large & 0.1 & \textbf{3.67\%} & \textbf{1.08\%}  & 3.42\% & 3.18\% \\
\hline
MNIST & Small & 0.3 & \textbf{43.10\%} & 14.87\%  & \textbf{33.64\%} & \textbf{33.64\%} \\
MNIST & Large & 0.3 & 45.66\% & \textbf{12.61\%}  & 41.62\% & 35.24\% \\
\hline
CIFAR10 & Small & 2/255 & 52.75\% & 38.91\%  & 39.35\% & 39.35\% \\
CIFAR10 & Large & 2/255 & 46.59\% & \textbf{31.28\%}  & 38.84\% & 36.08\% \\
CIFAR10 & Resnet & 2/255 & \textbf{46.11\%} & 31.72\%  & \textbf{36.41\%} & \textbf{35.93\%}\\
\hline
CIFAR10 & Small & 8/255 & 79.25\% & 72.24\%  & 71.71\% & 71.71\% \\
CIFAR10 & Large & 8/255 & 83.43\% & 80.56  & 79.24\% & 79.14\% \\
CIFAR10 & Resnet & 8/255 & \textbf{78.22\%} & \textbf{71.33\%}  & \textbf{70.95\%} & \textbf{70.77\%}\\
\end{tabular}
\label{fig:results}
\end{table}

\subparagraph{Summary of results} For the different data sets and models, the
final robust and nominal test errors are given in Table \ref{table:results}.  We
emphasize that in all cases we report the \emph{robust test 
error}, that is, our \emph{upper bound} on the possible test set error that the
classifier can suffer under \emph{any} norm-bounded attack (thus, considering
different empirical attacks is orthogonal to our main presentation and not
something that we include, as we are focused on verified performance).  As we
are focusing on the particular random projections discussed above, all
experiments consider attacks with bounded $\ell_\infty$ norm, plus the ReLU
networks highlighted above.  On MNIST, the (non-cascaded) large model reaches a
final robust error of 3.7\% for $\epsilon=0.1$, and the best cascade reaches
3.1\% error. This contrasts with the best previous bound of 5.8\% robust error
for this epsilon, from \citep{wong2017provable}. On CIFAR10, the ResNet model
achieves 46.1\% robust error for $\epsilon=2/255$, and the cascade
lowers this to 36.4\% error. In contrast, the previous best \emph{verified}
robust error for this $\epsilon$, from \citep{dvijotham2018dual}, was 80\%. 
While the robust error is naturally substantially higher for $\epsilon=8/255$ 
(the amount typically considered in empirical works), we are still able to
achieve 71\% provable robust error;  
for comparison, the best \emph{empirical}
robust performance against current attacks is 53\% error at $\epsilon=8/255$ 
\cite{madry2017towards}, and most heuristic defenses have been broken to
beyond this error \cite{athalye2018obfuscated}.


\begin{figure}
  \centering
  \includegraphics[scale=0.4]{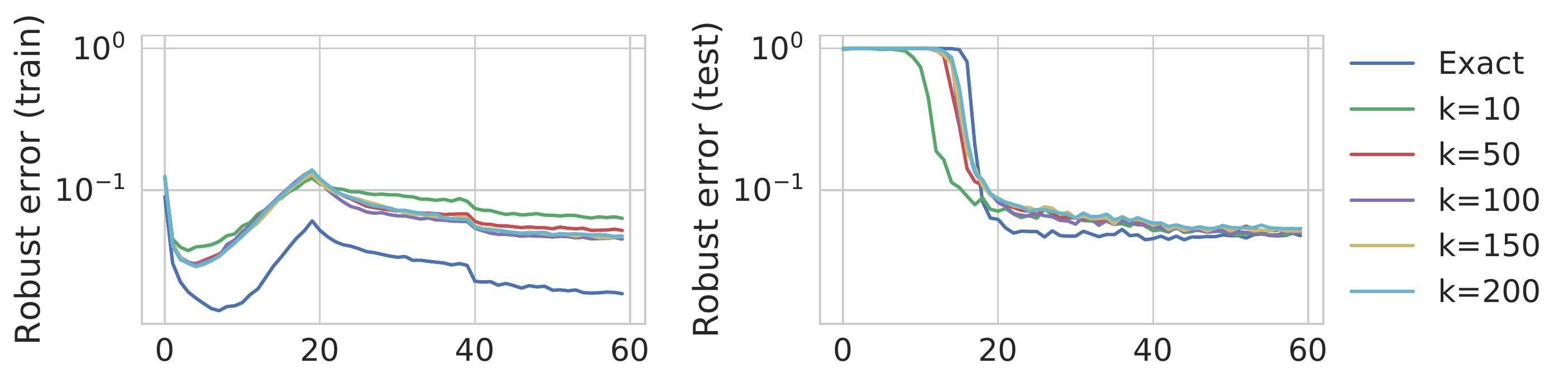}
  \caption{Training and testing robust error curves over epochs on the MNIST dataset using $k$ 
  projection dimensions. The $\epsilon$ value for training is scheduled from 0.01 to 0.1
  over the first 20 epochs. The projections force the model to generalize over higher variance, reducing the generalization gap.}
\label{fig:projvsexact}
\end{figure}

\subparagraph{Number of random projections}
In the MNIST dataset (the only data set where it is trivial to run exact
training without projection), we have evaluated our approach using different
projection dimensions as well as exact training (i.e., without random
projections). We note that using
substantially lower projection dimension does not
have a significant impact on the test error.  This fact is highlighted in
Figure \ref{fig:projvsexact}.  Using the same convolutional
architecture used by \citet{wong2017provable}, which previously required
gigabytes of memory and took hours to train, it is sufficient to use only 10
random projections to achieve comparable test error performance to training with
the exact bound.  Each training epoch with 10 random projections  
takes less than a minute on a single GeForce GTX 1080 Ti graphics card, while
using less than 700MB of memory, achieving significant speedup and 
memory reduction over \citet{wong2017provable}. The estimation quality and 
the corresponding speedups obtained are explored in more detail in 
Appendix \ref{app:cauchy}.

%

\begin{figure}
  \centering
  \includegraphics[scale=0.4]{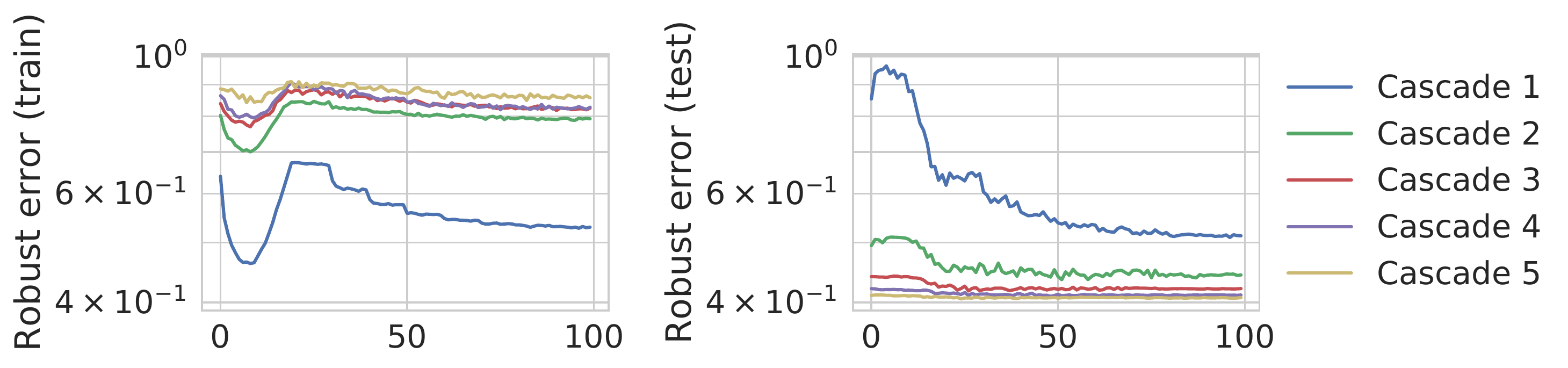}
  \caption{Robust error curves as we add models to the cascade for the CIFAR10 dataset on a small model. The $\epsilon$ value for training is scheduled to reach 2/255 after 20 epochs. The training curves are for each individual model, and the testing curves are
  for the whole cascade up to the stage. }
\label{fig:cifar}
\end{figure}

\subparagraph{Cascades}
Finally, we consider the performance of the cascaded versus non-cascaded models.
In all cases, cascading the models is able to improve the robust error
performance, sometimes substantially, for instance decreasing the robust error
on CIFAR10 from 46.1\% to 36.4\% for $\epsilon=2/255$.  However, this comes at a cost as
well: the \emph{nominal} error \emph{increases} throughout the cascade (this is
to be expected, since the cascade essentially tries to force the robust and
nominal errors to match).  Thus, there is substantial value to both improving
the single-model networks \emph{and} integrating cascades into the prediction.

\section{Conclusion}
In this paper, we have presented a general methodology for deriving 
dual networks from compositions of dual layers based on the 
methodology of conjugate functions to train classifiers 
that are provably robust to adversarial attacks. 
Importantly, the methodology is linearly scalable for ReLU based networks 
against $\ell_\infty$ norm bounded attacks, making it possible 
to train large scale, provably robust networks that 
were previously out of reach, and the obtained bounds 
can be improved further with model cascades. 
While this marks a significant step forward in scalable defenses 
for deep networks, there are several directions for improvement. One
particularly important direction is better architecture development:
a wide range of functions and activations not found in traditional deep residual
networks may have better robustness properties or more efficient dual layers
that also allow for scalable training.  But perhaps even more importantly, we
also need to consider the nature of adversarial perturbations beyond just
norm-bounded attacks.  Better characterizing the space of perturbations that a
network ``should'' be resilient to represents one of the major challenges going
forward for adversarial machine learning.

\bibliography{convex_adversarial}
\bibliographystyle{plainnat}

\clearpage
\appendix
\section{Conjugates and lower bounds with duality}
\subsection{Conjugates of the joint indicator function}
\label{app:conjugates} 
Here, we derive a lower bound on $\sum_{i=2}^k \chi_{i}(z_{1:i})$. It is mathematically convenient to introduce addition variables $\hat z_{1:k}$ such that $\hat z_i = z_i$ for all $i=1,\dots, k$, and rephrase it as the equivalent constrained optimization problem. 
\begin{equation}
\begin{split}
& \min_{z_{1:k-1}, \hat z_{2:k}} 0\\
\subjectto \hat z_i &= \sum_{j=1}^{i-1} f_{ij}(z_j)\;\;\for \;\; i=2,\dots, k \\
z_i &= \hat z_i \;\; \for i=1, \dots, k
\end{split}
\end{equation}
Note that we do not optimize over $\hat z_1$ and $z_k$ yet, to allow for future terms 
on the inputs and outputs of the network, so this is analyzing just the network structure. 
We introduce Lagrangian variables $\nu_{1:k}, \hat \nu_{2:k}$ to get the following Lagrangian: 
\begin{equation}
L(z_{1:k}, \hat z_{1:k}, \nu_{1:k}, \hat \nu_{2:k}) = \sum_{i=2}^k \hat \nu_i^T \left(\hat z_i - \sum_{j=1}^{i-1} f_{ij}(z_j)\right) + \sum_{i=1}^k\nu_i^T \left(z_i - \hat z_i\right)
\end{equation}
Grouping up terms by $z_i, \hat z_i$ and rearranging the double sum results in the following expression:
\begin{equation}
L(z_{1:k}, \hat z_{1:k}, \nu_{1:k}, \hat \nu_{2:k}) = - \nu_1^T\hat z_1 + \sum_{i=2}^k (\hat \nu_i -\nu_i)^T\hat z_i + \sum_{i=1}^{k} \left(\nu_i^T z_i - \sum_{j=i+1}^k \hat \nu_j^Tf_{ji}(z_i)\right)
\end{equation}
From the KKT stationarity conditions for the derivative with respect to $\hat z_i$, we know that $\hat \nu_i = \nu_i$. Also note that in the summand, the last term for $i=k$  has no double summand, so we move it out for clarity. 
\begin{equation}
L(z_{1:k}, \nu_{1:k}) = - \nu_1^T\hat z_1 + \nu_k^Tz_k + \sum_{i=1}^{k-1} \left(\nu_i^T z_i - \sum_{j=i+1}^k \nu_j^Tf_{ji}(z_i)\right)
\end{equation}
Finally, we minimize over $z_i$ for $i=2, \dots, k-1$ to get the conjugate form for the lower bound via weak duality.
\begin{equation}
\begin{split}
L(z_{1:k}, \nu_{1:k}) &\geq - \nu_1^T\hat z_1 + \nu_k^Tz_k + \sum_{i=1}^{k-1} \min_{z_i}\left(\nu_i^T z_i - \sum_{j=i+1}^k \nu_j^Tf_{ji}(z_i)\right)\\
&=- \nu_1^T\hat z_1 + \nu_k^Tz_k - \sum_{i=1}^{k-1} \max_{z_i}\left(-\nu_i^T z_i + \sum_{j=i+1}^k \nu_j^Tf_{ji}(z_i)\right)\\
&= - \nu_1^Tz_1 + \nu_k^Tz_k - \sum_{i=1}^{k-1} \chi_i^*(-\nu_i, \nu_{i+1:k})
\end{split}
\end{equation}


\subsection{Proof of Theorem \ref{thm:dual_network}}
\label{app:thm1}
First, we rewrite the primal problem by bringing the function and input constraints into the objective with indicator functions $I$. We can then apply Lemma \ref{cor:support} to get the 
following lower bound on the adversarial problem: 
\begin{equation}
  \begin{split}
    \maximize_{\nu_{1:k}}\minimize_{z_1, z_k} \;\; & (c^T+ \nu_k)^Tz_k + I_{\mathcal B(x)}(z_1) - \nu_1^Tz_1 - \sum_{i=1}^{k-1} \chi_i^*(-\nu_i, \nu_{i+1:k})
    \end{split}
\end{equation}
Minimizing over $z_1$ and $z_k$, note that
\begin{equation}
\begin{split}
& \min_{\hat z_k} (c+\nu_k)^T\hat z_k = -I(\nu_k = -c)\\
& \min_{\hat z_1} I_{\mathcal B(x)}(z_1) - \nu_1^T z_1 = -I^*_{\mathcal B(x)}(\nu_1)\\
\end{split}
\end{equation}
Note that if $\mathcal B(x) = \{x +\Delta : \|\Delta\| \leq \epsilon\}$ for some norm, then $I^*_{\mathcal B(x)}(\nu_1) =\nu_1^Tx + \epsilon\|\nu_1\|_*$ where $\|\cdot\|$ is the dual norm, but any sort of input constraint can be used so long as its conjugate can be bounded. Finally, the last term can be bounded with the dual layer: 
\begin{equation}
\begin{split}
&\min_{z_i}\nu_i^T z_i - \sum_{j={i+1}}^{k} \nu_{j}^Tf_{ji}(z_i) = -\chi_i^*(-\nu_i,\nu_{i+1:k})
\geq -h_i(\nu_{i:k})\;\; \subjectto\;\; \nu_i = \sum_{j=i+1}^k g_{ij}(\nu_j)
\end{split}
\end{equation}
Combining these all together, we get that the adversarial problem from Equation \ref{eq:primal} is lower bounded by 
\begin{equation}
  \begin{split}
    \maximize_{\nu} \;\; & -\nu_1^Tx - \epsilon\|\nu_1\|_* - \sum_{i=1}^{k-1}h_i(\nu_{i:k})\\
    \subjectto \;\; \nu_k &= -c\\
    \nu_i &= \sum_{j=i+1}^k g_{ij}(\nu_j)
    \end{split}
\end{equation}

\section{Dual layers}
\label{app:dual_layers}
In this section, we derive the dual layers for standard building blocks of deep learning. 
\subsection{Linear operators}
Suppose $f_i(z_i) = W_iz_i + b_i$ for some linear operator $W_i$ and bias terms $b_i$. Then, 
\begin{equation}
\begin{aligned}
\chi_i^*(-\nu_i, \nu_{i+1}) &= \max_{z_i} -z_i^T\nu_i + (W_iz_i + b_i)^T\nu_{i+1}\\
&= \max_{z_i} z_i^T(W_i^T\nu_{i+1} - \nu_i) + b_i^T\nu_{i+1}\\
&= \max_{z_i} I(\nu_i = W_i^T\nu_{i+1}) + b_i^T\nu_{i+1}\\
&= b_i^T\nu_{i+1}\;\; \subjectto \;\; \nu_i = W_i^T\nu_{i+1}
\end{aligned}
\end{equation}

\subsection{Residual linear connections}
Suppose $f_i(z_i, z_j) = W_iz_i + z_j + b_i$ and $z_{j+1} = W_jz_j + b_j$ for some $j < i-1$ for linear operators $W_i, W_j$ and bias term $b_i, b_j$. Then, 
\begin{equation}
\begin{aligned}
\chi_i^*(-\nu_i, \nu_{i+1}) &= \max_{z_i} -z_i^T\nu_i + (W_iz_i + b_i)^T\nu_{i+1}\\
&= b_i^T\nu_{i+1}\;\; \subjectto \;\; \nu_i = W_i^T\nu_{i+1}
\end{aligned}
\end{equation}
and
\begin{equation}
\begin{aligned}
\chi_i^*(-\nu_j, \nu_{j+1}) &= \max_{z_j} -z_j^T\nu_j + z_j^T\nu_i + (W_jz_j + b_j)^T\nu_{j+1}\\
&= b_j^T\nu_{j}\;\; \subjectto \;\; \nu_j = W_j^T\nu_{j+1} + \nu_i
\end{aligned}
\end{equation}

\subsection{ReLU activations}
The proof here is the same as that presented in Appendix A3 of \citet{wong2017provable}, 
however we reproduce a simplified version here for the reader. The conjugate function
for the ReLU activation is the following: 
\begin{equation}
\chi^*(-\nu_i, \nu_{i+1}) = \max_{z_i} -z_i^T\nu_i + \max(z_i,0)\nu_{i+1}
\end{equation}
Suppose we have lower and upper bounds $\ell_i, u_i$ on the input $z_i$. If $u_i \leq 0$, then $\max(z_i,0) = 0$, and so 
\begin{equation}
\chi^*(-\nu_i, \nu_{i+1}) = \max_{z_i} -z_i^T\nu_i = 0\;\; \subjectto\;\; \nu_i = 0
\end{equation}
Otherwise, if $\ell_i \geq 0$, then $\max(z_i,0) = z_i$ and we have
\begin{equation}
\chi^*(-\nu_i, \nu_{i+1}) = \max_{z_i} -z_i^T\nu_i + z_i^T\nu_{i+1} = 0\;\; \subjectto\;\; \nu_i = \nu_{i+1}
\end{equation}
Lastly, suppose $\ell_i < 0 < u_i$. Then, we can upper bound the conjugate by taking the maximum 
over a convex outer bound of the ReLU, namely $\mathcal S_i = \{(z_i, z_{i+1}) : z_{i+1} \geq 0, z_{i+1} \geq z_i, -u_i\odot z_{i} + (u_i-\ell_i)\odot z_{i+1} \leq -u_i\odot\ell_i\}$, where $\odot$ denotes element-wise multiplication: 
\begin{equation}
\chi^*(-\nu_i, \nu_{i+1}) \leq \max_{\mathcal S_i} -z_i^T\nu_i + z_{i+1}^T\nu_{i+1}
\end{equation}
The maximum must occur either at the origin $(0,0)$ or along the line $-u_{ij}z_{ij} + (u_{ij}-\ell_{ij}) z_{i+1,j} = -u_{ij}\ell_{ij}$, so we can upper bound it again with 
\begin{equation}
\begin{split}
\chi^*(-\nu_{ij}, \nu_{i+1,j}) &\leq  \max_{z_{ij}} \left[-z_{ij}\nu_{ij} + \left( \frac{u_{ij}}{u_{ij}-\ell_{ij}}z_{ij}-\frac{u_{ij}\ell_{ij}}{u_{ij}-\ell_{ij}}\right)\nu_{i+1,j}\right]_+\\
&=  \max_{z_{ij}} \left[\left(\frac{u_{ij}}{u_{ij}-\ell_{ij}}\nu_{i+1,j}-\nu_{ij}\right)z_{ij}-\frac{u_{ij}\ell_{ij}}{u_{ij}-\ell_{ij}}\nu_{i+1,j}\right]_+\\
&= \left[-\frac{u_{ij}\ell_{ij}}{u_{ij}-\ell_{ij}}\nu_{i+1,j}\right]_+ \;\; \subjectto \;\; \nu_{ij} = \frac{u_{ij}}{u_{ij}-\ell_{ij}} \nu_{i+1,j}\\
&= -\ell_{ij}\left[\nu_{ij}\right]_+ \;\; \subjectto \;\; \nu_{ij} = \frac{u_{ij}}{u_{ij}-\ell_{ij}} \nu_{i+1,j}
\end{split}
\end{equation}
Let $\mathcal I_i^-, \mathcal I_i^+, \mathcal I$ and $D_i$ be as defined in the corollary. Combining these three cases together, we get the final upper bound: 
\begin{equation}
\chi_{i}^*(-\nu_i, \nu_{i+1:k}) \leq - \sum_{j\in \mathcal I_i} \ell_{i,j} [\nu_{i,j}]_+ \;\; \subjectto \;\; \nu_i = D_i\nu_{i+1}
\end{equation}

%

\subsection{Hardtanh}
Here, we derive a dual layer for the hardtanh activation function. 
The hard tanh activation function is given by
\begin{equation}
\hardtanh(x) = \begin{cases}
-1 &\for\;\; x < -1\\
x &\for\;\; -1 \leq x \leq 1\\
1 &\for\;\; x > 1
\end{cases}
\end{equation}

Since this is an activation function (and has no skip connections), we only need to bound the following: 
\begin{equation}
\chi^*(-\nu_i, \nu_{i+1}) = \max_{z_i} -z_i^T\nu_i + \hardtanh(z_i)^T\nu_{i+1}
\end{equation}
Given lower and upper bounds $\ell$ and $u$, we can use a similar convex relaxation as that used for ReLU and decompose this problem element-wise (we will now assume all terms are scalars for notational simplicity), so we have 
\begin{equation}
\chi^*(\nu_i, \nu_{i+1})\leq \max_{z_i, z_{i+1} \in \mathcal S} -z_i\nu_i + z_{i+1}\nu_{i+1}
\end{equation}
where $\mathcal S$ is the convex relaxation. The exact form of the relaxation depends on the values of $\ell$ and $u$, and we proceed to derive the dual layer for each case. We depict the relaxation where $u > 1$ and $\ell < -1$ in Figure \ref{fig:hardtanh}, and note that the remaining cases are either triangular relaxations similar to the ReLU case or exact linear regions. 

\subsubsection{$u > 1, \ell < -1$}

\begin{figure}
\centering
\includegraphics[scale=0.5]{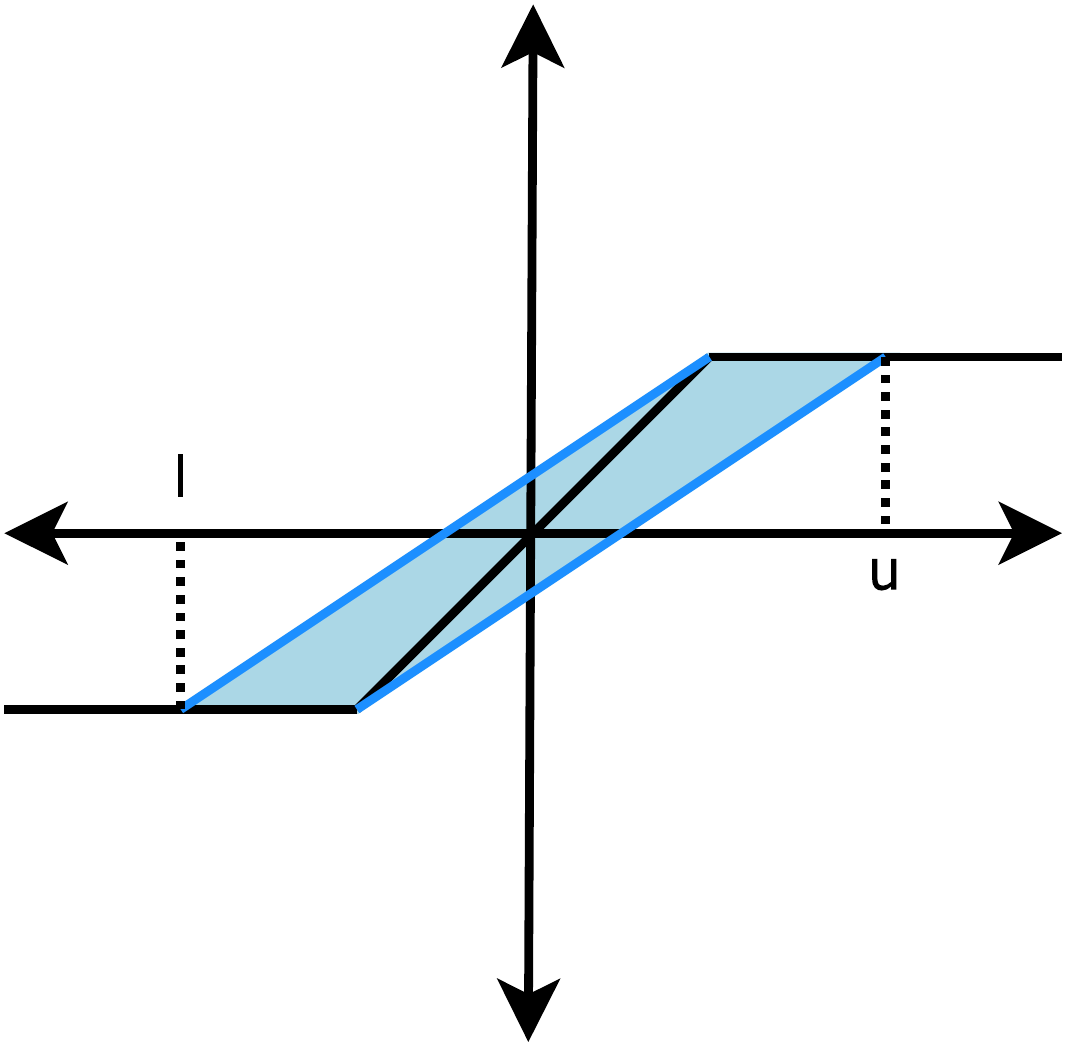}
\caption{Convex relaxation of hardtanh given lower and upper bounds $\ell$ and $u$.}
\label{fig:hardtanh}
\end{figure}

If $u> 1$ and $\ell<-1$, we can use the relaxation given in Figure \ref{fig:hardtanh}.
The upper bound goes through the points $(\ell, -1)$ and $(1,1)$ while the lower bound goes through the points $(-1,-1)$ and $(u, 1)$. The slope of the first one is $\frac{2}{1-\ell}$ and the slope of the second one is $\frac{2}{u+1}$, so we have either
\begin{equation}
z_{i+1} = \frac{2}{1-\ell}(z_i-1) + 1, \;\; z_{i+1} = \frac{2}{u+1}(z_i+1)-1
\end{equation}
Taking the maximum over these two cases, we have our upper bound of the conjugate is 
\begin{equation}
\chi^*(\nu_i, \nu_{i+1})\leq \max\left(-z_i\nu_i + \left(\frac{2}{1-\ell}(z_i-1) + 1\right)\nu_{i+1}, -z_i\nu_i + \left( \frac{2}{u+1}(z_i+1)-1\right)\nu_{i+1}\right)
\end{equation}
Simplifying we get
\begin{equation}
\begin{split}
\chi^*(\nu_i, \nu_{i+1})\leq \max&\left(z_i\left(-\nu_i + \frac{2}{1-\ell}\nu_{i+1}\right)+\left(1-\frac{2}{1-\ell}\right)\nu_{i+1},\right.\\
&\left. z_i\left(-\nu_i + \frac{2}{u+1}\nu_{i+1}\right)+\left(\frac{2}{u+1} - 1\right)\nu_{i+1}\right)
\end{split}
\end{equation}
So each case becomes
\begin{equation}
\begin{split}
\chi^*(\nu_i, \nu_{i+1})\leq \max&\left(\left(1-\frac{2}{1-\ell}\right)\nu_{i+1}\;\;\subjectto\;\; \nu_i = \frac{2}{1-\ell}\nu_{i+1}\right., \\
&\;\;\left.\left(\frac{2}{u+1} - 1\right)\nu_{i+1}\;\; \subjectto\;\;\nu_i = \frac{2}{u+1}\nu_{i+1}\right)
\end{split}
\end{equation}
As a special case, note that when $u=-\ell$, we have 
\begin{equation}
\chi^*(\nu_i, \nu_{i+1})\leq \left|\left(1 - \frac{2}{1+u}\right)\nu_{i+1}\right|\;\;\subjectto\;\; \nu_i = \frac{2}{1+u}\nu_{i+1}
\end{equation}
This dual layer is linear, and so we can continue to use random projections for efficient bound estimation. 

\subsubsection{$u \leq -1$}
Then, $\mathcal S = \{z_{i+1} = -1\}$ and so 
\begin{equation}
\chi^*(\nu_i, \nu_{i+1}) = \max_{z_i} -z_i\nu_i - \nu_{i+1} = - \nu_{i+1} \;\;\subjectto \;\; \nu_i = 0
\end{equation}
\subsubsection{$\ell \geq 1$}
Then, $\mathcal S = \{z_{i+1} = 1\}$ and so 
\begin{equation}
\chi^*(\nu_i, \nu_{i+1}) = \max_{z_i} -z_i\nu_i + \nu_{i+1} = \nu_{i+1} \;\;\subjectto \;\; \nu_i = 0
\end{equation}
\subsubsection{$\ell \geq -1, u \leq 1$}
Then, $\mathcal S = \{z_{i+1} = z_i\}$ and so 
\begin{equation}
\chi^*(\nu_i, \nu_{i+1}) = \max_{z_i} -z_i\nu_i + z_i\nu_{i+1} = 0 \;\;\subjectto \;\; \nu_i = \nu_{i+1}
\end{equation}
\subsubsection{$\ell \leq -1, -1 \leq u \leq 1$}
Here, our relaxation consists of the triangle above the hardtanh function. 
Then, the maximum occurs either on the line $z_{i+1} = \frac{1+u}{u-\ell}(z_i - \ell) -1$ or  at $(-1,-1)$. This line is equivalent to $z_{i+1} = \frac{1+u}{u-\ell}z_i - \left(\frac{1+u}{u-\ell}\ell + 1\right)$, and the point $(-1,-1)$ has objective value $\nu_i - \nu_{i+1}$, so we get
\begin{equation}
\chi^*(\nu_i, \nu_{i+1}) \leq \max_{z_{i}} -z_i\nu_i + \frac{1+u}{u-\ell}z_i\nu_{i+1} - \left(\frac{1+u}{u-\ell}\ell + 1\right)\nu_{i+1}
\end{equation}
\begin{equation}
\chi^*(\nu_i, \nu_{i+1}) \leq \max\left(-\left(\frac{1+u}{u-\ell}\ell + 1\right)\nu_{i+1}, \nu_i - \nu_{i+1} \right) \;\;\subjectto \;\; \nu_i = \frac{1+u}{u-\ell} \nu_{i+1}
\end{equation}
\subsubsection{$-1 \leq \ell \leq 1, 1 \leq u$}
Here, our relaxation consists of the triangle below the hardtanh function. 
Then, the maximum occurs either on the line $z_{i+1} = \frac{1-\ell}{u-\ell}(z_i - \ell) +\ell$ or  at $(1,1)$. This line is equivalent to $z_{i+1} =  \frac{1-\ell}{u-\ell}z_i - \left(\frac{1-\ell}{u-\ell}\ell - \ell\right)$, and at the point $(1,1)$ has objective value $-\nu_i + \nu_{i+1}$, so we get

\begin{equation}
\chi^*(\nu_i, \nu_{i+1}) \leq \max_{z_{i}} -z_i\nu_i + \frac{1-\ell}{u-\ell}z_i\nu_{i+1} - \left(\frac{1-\ell}{u-\ell}\ell - \ell\right)\nu_{i+1}
\end{equation}
\begin{equation}
\chi^*(\nu_i, \nu_{i+1}) \leq \max\left(-\left(\frac{1-\ell}{u-\ell}\ell - \ell\right)\nu_{i+1}, -\nu_i + \nu_{i+1} \right) \;\;\subjectto \;\; \nu_i =  \frac{1-\ell}{u-\ell} \nu_{i+1}
\end{equation}

\subsection{Batch normalization}
As mentioned before, we only consider the case of batch normalization with a fixed mean and variance. 
This is true during test time, and at training time 
we can use the batch statistics as a heuristic. 
Let $\mu_i,\sigma_i$ be the fixed mean and variance statistics, so batch
normalization has the following form: 
\begin{equation}
BN(z_i) = \gamma\frac{x_i- \mu_i}{\sqrt{\sigma_i^2 + \epsilon}} + \beta
\end{equation}
where $\gamma,\beta$ are the batch normalization parameters. Then, 
\begin{equation}
\label{eq:batch_form}
  \begin{split}
    z_{i} &= \gamma\frac{\hat{z}_{i} - \mu}{\sqrt{\sigma^2 + \epsilon}} + \beta =D_{i}z_{i} + d_{i}\\ 
  \end{split}
\end{equation}
where $D_{i+1} = \diag\left(\frac{\gamma}{\sqrt{\sigma^2 + \epsilon}}\right)$ and $d_{i+1} = \beta - \frac{\mu}{\sqrt{\sigma^2 + \epsilon}}$. 
and so we can simply plug this into the linear case to get 
\begin{equation}
\chi_{i}^*(-\nu_i, \nu_{i+1:k}) = d_i^T\nu_{i+1}   \;\; \subjectto \;\; \nu_i = D_i\nu_{i+1}
\end{equation}
Note however, that batch normalization has the effect of shifting the activations to be 
centered more around the origin, which is exactly the case in which the robust 
bound becomes looser. In practice, we find that while including batch normalization 
may improve convergence, it reduces the quality of the bound. 

\section{Cascade construction}

\begin{algorithm}[tb]
   \caption{Training robust cascade of $k$ networks and making predictions}
   \label{alg:cascade}
\begin{algorithmic}
  \STATE \textbf{input:} Initialized networks $f_1,\dots, f_k$, training examples $X,y$, robust training procedure denoted RobustTrain, test example $x^*$
  \FOR{$i=1,\ldots,k$}
  \STATE $f_i$ := RobustTrain$(f_i,X,y)$ \emph{// Train network}
  \STATE \emph{// remove certified examples from dataset}
  \STATE $X,y := \{x_i,y_i : J(x,g(e_{f(x_i)} - e_{y^{targ}})) > 0, \;\; \forall y^{targ} \neq f(x_i)\}$
  \ENDFOR
  \FOR{$i=1,\dots,k$}
  \IF{$J(x,g(e_{f_i(x^*)} - e_{y^{targ}})) < 0 \;\; \forall y^{targ} \neq f_i(x^*)$}
  \STATE \textbf{output:} $f_i(x^*)$ \emph{// return label if certified}
  \ENDIF
  \ENDFOR
  \STATE \textbf{output:} no certificate
\end{algorithmic}
\end{algorithm}

\begin{figure}
  \centering
  \includegraphics[scale=0.5]{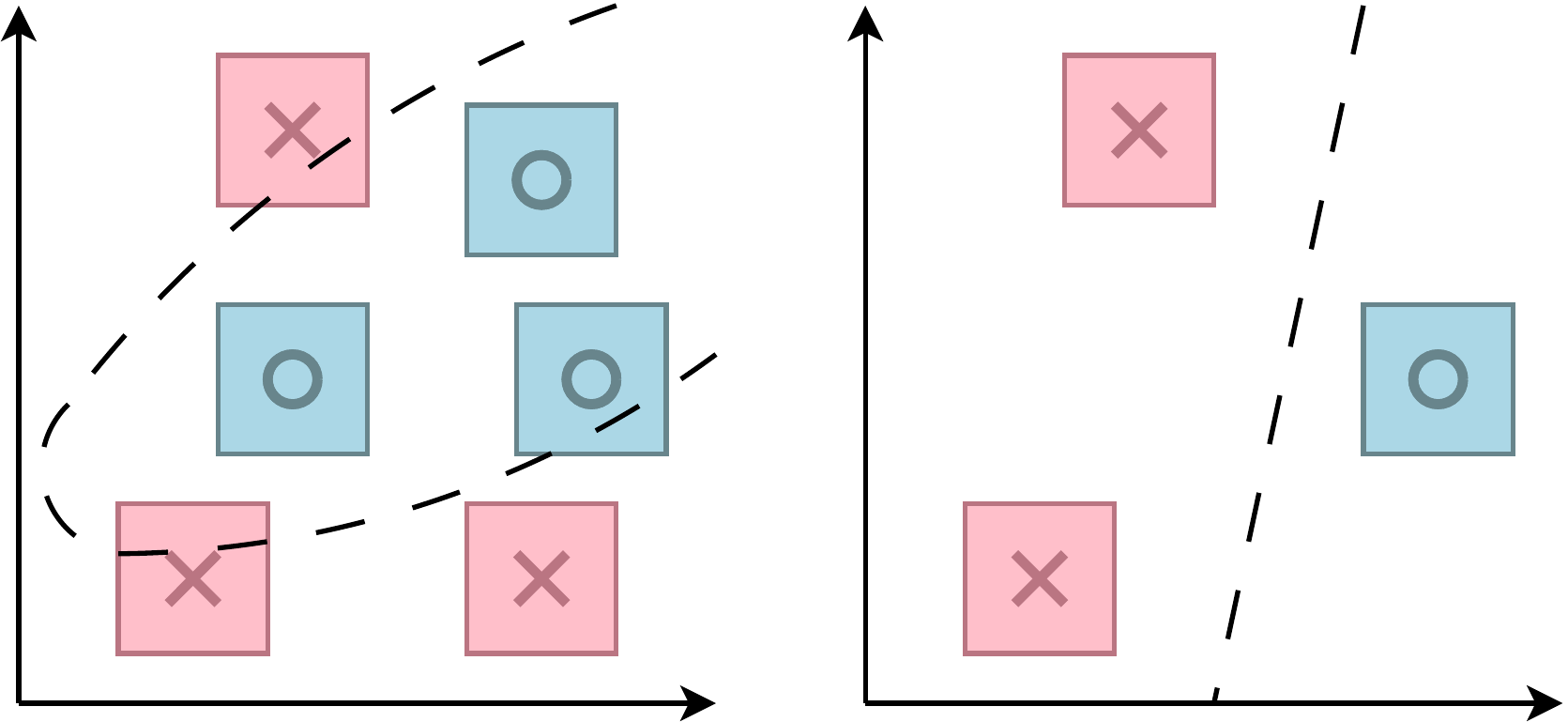}
  \caption{An example of a two stage cascade. The first model on the left can only robustly classify three of the datapoints. After removing the certified examples, the remaining examples can now easily be robustly classified by a second stage classifier. }
\label{fig:cascade}
\end{figure}

The full algorithm for constructing cascades as we describe in the main text is
shown in Algorithm \ref{alg:cascade}.
To illustrate the use of the cascade, Figure \ref{fig:cascade} shows a two stage cascade
on a few data points in two dimensional space. The boxes denote the adversarial ball around
each example, and if the decision boundary is outside of the box, the example is certified. 

\section{Estimation using Cauchy random projections}
\subsection{Proof of Theorem \ref{thm:projection}} 
\label{app:thm2}
\paragraph{Estimating $\|\hat{\nu}_1\|_{1,:}$}
Recall the form of $\hat{\nu}_1$, 
$$\hat{\nu}_1 = IW_1^T D_2 W_2^T \ldots D_n W_n^T = g(I)$$
where we include the identity term to make explicit the fact that we compute this
by passing an identity matrix through the network $g$. 
Estimating this term is straightforward: we simply pass in a Cauchy random matrix $R$, 
and take the median absolute value: 
$$\|\hat{\nu}_1\|_{1,:} \approx \median(|RW_1^T D_2 W_2^T \ldots D_n W_n^T|) = \median(|g(R)|)$$
where the median is taken over the minibatch axis. 
\paragraph{Estimating $\sum_{i} [\nu_{i,:}]_+$}
Recall the form of $\nu = \nu_j$ for some layer $j$, 
$$\nu_j = ID_j W_j^T \ldots D_n W_n^T = g_j(I)$$
Note that for a vector $x$,  
$$\sum_i [x]_+ = \frac{\|x\|_1 + 1^Tx}{2}$$
So we can reuse the $\ell_1$ approximation from before to get
$$\sum_i [\nu_{i,:}]_+ = \frac{\|\nu\|_{1,_:} + 1^T\nu}{2} \approx \frac{|\median(g_j(R)) + g_j(1^T)|}{2} $$
which involves using the same median estimator and also passing in a single example of ones through the network. 
\paragraph{Estimating $\sum_{i \in \mathcal{I}} \ell_i [\nu_{i,:}]_+$}
The previous equation, while simple, is not exactly the term in the objective; 
there is an addition $\ell_1$ factor for each row, and we only add rows in the 
$\mathcal{I}$ set. However, we can deal with this by simply passing in a 
modified input to the network, as we will see shortly: 
\begin{equation}
\begin{aligned}
\sum_{i \in \mathcal{I}} \ell_i [\nu_{i,:}]_+ &= \sum_{i \in \mathcal{I}} \ell_i \frac{|\nu_{i,:}| + \nu_{i,:}}{2}\\
&= \frac{1}{2}\left(\sum_{i \in \mathcal{I}} \ell_i |\nu_{i,:}| + \sum_{i \in \mathcal{I}} \ell_i\nu_{i,:}\right)\\
&= \frac{1}{2}\left(\sum_{i \in \mathcal{I}} \ell_i |g_j(I)_i| + \sum_{i \in \mathcal{I}} \ell_ig_j(I)_i\right)
\end{aligned}
\end{equation}
Note that since $g_j$ is just a linear function that does a forward pass through the network,
for any matrix $A,B$, 
$$Ag_j(B) = ABD_jW_j^T\ldots D_n W_n^T =  g_j(AB).$$ So we can take the multiplication by 
scaling terms $\ell$ to be an operation on the input to the network (note that we assume $\ell_i < 0$, which is true for all $i \in \mathcal{I}$)
\begin{equation}
\begin{aligned}
\sum_{i \in \mathcal{I}} \ell_i [\nu_{i,:}]_+ &= \frac{1}{2}\left(-\sum_{i \in \mathcal{I}} |g_j(\diag(\ell))_i| + \sum_{i \in \mathcal{I}} g_j(\diag(\ell))_i\right)
\end{aligned}
\end{equation}
Similarly, we can view the summation over the index set $\mathcal{I}$ as a summation
after multiplying by
an indicator matrix $1_\mathcal{I}$ which zeros out the ignored rows. Since
this is also linear, we can move
it to be an operation on the input to the network.
\begin{equation}
\begin{aligned}
\sum_{i \in \mathcal{I}} \ell_i [\nu_{i,:}]_+ &= \frac{1}{2}\left(-\sum_{i} |g_j(1_\mathcal{I}\diag(\ell))_i| + \sum_{i} g_j(1_\mathcal{I}\diag(\ell))_i\right)
\end{aligned}
\end{equation}
Let the linear, preprocessing operation be $h(X) = X 1_\mathcal{I}\diag(\ell)$ so 
$$h(I) = 1_\mathcal{I}\diag(\ell).$$ 
Then,
we can observe that the two terms are simply an $\ell_{1,:}$ operation and a summation of the network output after applying $g_j$ to $h(I)$ (where in the latter case, since
everything is linear we can take the summation inside both $g$ and $h$ to make it $g_j(h(1^T))$):
\begin{equation}
\sum_{i \in \mathcal{I}} \ell_i [\nu_{i,:}]_+ = \frac{1}{2}\left(- \|g_j(h(I))\|_{1,:} + g_j(h(1^T))\right)
\end{equation}
The latter term is cheap to compute, since we only pass a single vector. We can approximate the first term using the median estimator on the compound operations $g\circ h$ for a Cauchy random matrix $R$: 
\begin{equation}
\sum_{i \in \mathcal{I}} \ell_i [\nu_{i,:}]_+ \approx \frac{1}{2}\left(- \median(|g_j(h(R))|) + g_j(h(1^T))\right)
\end{equation}
The end result is that this term can be estimated by generating a Cauchy random matrix,
 scaling its terms by $\ell$ and zeroing out columns in $\mathcal{I}$, then passing it through the network and taking the median. $h(R)$ can be computed for each layer lower bounds $\ell$, and cached to be computed for the next layer, similar to the non-approximate case. 
 
\section{High probability bounds}
In this section, we derive high probability certificates for robustness against adversarial examples. Recall that the original certificate is of the form
$$J(g(c,\alpha)) < 0,$$
so if this holds we are guaranteed that the example cannot be adversarial. 
What we will show is an equivalent high probability statement: for $\delta > 0$, with probability at least $(1-\delta)$, 
$$J(g(c,\alpha)) \leq \tilde J(g(c,\alpha))$$
where $\tilde J$ is equivalent to the original $J$ but using a high probability $\ell_1$ upper bound. Then, if $\tilde J(g(c,\alpha)) <0$ then with high probability we have a certificate.

\subsection{High probability bounds using the geometric estimator}
\label{app:high_prob}
While the median estimator is a good heuristic for training, it is 
still only an estimate of the bound. At test time, it is possible to create a provable bound 
that holds with high probability, which may be desired if computing the 
exact bound is computationally impossible. 

In this section, we derive high probability certificates for robustness against adversarial examples. Recall that the original certificate is of the form
$$J(g(c,\alpha)) < 0,$$
so if this holds we are guaranteed that the example cannot be adversarial. 
What we will show is an equivalent high probability statement: for $\delta > 0$, with probability at least $(1-\delta)$, 
$$J(g(c,\alpha)) \leq \tilde J(g(c,\alpha))$$
where $\tilde J$ is equivalent to the original $J$ but using a high probability upper bound
on the $\ell_1$ norm. Then, if $\tilde J(g(c,\alpha)) <0$ then with high probability we have a certificate. 

\subsection{Tail bounds for the geometric estimator}
From \citet{li2007nonlinear}, the authors also provide a geometric mean 
estimator which comes with
high probability tail bounds. The geometric estimator is 
$$\|\hat{\nu}_1\|_{1,j} \approx \prod_{i=1}^k |g(R)_{i,j}|^{1/k}$$
and the relevant lower tail bound on the $\ell_1$ norm is 
\begin{equation}
\label{eq:prob_bound}
P\left(\frac{1}{1-\epsilon}\prod_{i=1}^k |g(R)_{i,j}|^{1/k} \leq \|\hat{\nu}_1\|_{1,j}\right) \leq \exp\left(-k\frac{\epsilon^2}{G_{L,gm}}\right)
\end{equation}
where 
$$G_{L,gm} = \frac{\epsilon^2}{\left(-\frac{1}{2}\log\left(1 + \left(\frac{2}{\pi}\log(1-\epsilon)\right)^2\right) + \frac{2}{\pi}\tan^{-1}\left(\frac{2}{\pi}\log(1-\epsilon)\right)\log(1-\epsilon)\right)}$$
Thus, if $\exp\left(-k\frac{\epsilon^2}{G_{L,gm}}\right) \leq \delta$, then with probability $1-\delta$ we have that 
$$\|\hat{\nu}_1\|_{1,j} \leq \frac{1}{1-\epsilon}\prod_{i=1}^k |g(R)_{i,j}|^{1/k} = \textrm{geo}(R)$$
which is a high probability upper bound on the $\ell_1$ norm. 

\subsection{Upper bound on $J(g(c,\alpha))$}
In order to upper bound $J(g(c,\alpha))$, we must apply the $\ell_1$ upper bound for \emph{every} $\ell_1$ term. Let $n_1, \dots, n_k$ denote the number of units in each layer of a $k$ layer neural network, then we enumerate all estimations as follows: 
\begin{enumerate}
\item The $\ell_1$ norm computed at each intermediary layer when computing iterative bounds. This results in $n_2 + \dots + n_{k-1}$ estimations. 
\item The $\sum_{j\in \mathcal{I}_i} \ell_{i,j}[\nu_{i,j}]_+$ term for each $i=2, \dots, k-1$, computed at each intermediary layer when computing the bounds. This results in $n_3 + 2n_4 + \dots + (k-3)n_{k-1}$. 
\end{enumerate}
In total, this is $n_2 + 2n_3 + \dots + (k-2)n_{k-1} = N$ total estimations. In order to say that \emph{all} of these estimates hold with probability $1-\delta$, we can do the following: we bound each estimate in Equation \ref{eq:prob_bound} with probability $\delta/N$, and use the union bound over all $N$ estimates. We can then conclude that with probability at most $\delta$, any estimate is not an upper bound, and so with probability $1-\delta$ we have a proper upper bound. 

\subsection{Achieving $\delta/N$ tail probability}
There is a problem here: if $\delta/N$ is small, then $\epsilon$ becomes large, and the bound gets worse. In fact, since $\epsilon<1$, when $k$ is fixed, there's actually a lower limit to how small $\delta/N$ can be. 

To overcome this problem, we take multiple samples to reduce the probability. Specifically, 
instead of directly using the geometric estimator, we use the maximum over multiple
geometric estimators 
$$\textrm{maxgeo}(R_1, \dots, R_m) = \max(\textrm{geo}(R_1), \dots, \textrm{geo}(R_m)),$$ 
where $R_i$ are independent Cauchy random matrices. If each one has a tail probability of $\delta$, then the maximum has a tail probability of $\delta^m$, which allows us to get arbitrarily small tail probabilities at a rate exponential in $m$. 

\subsection{High probability tail bounds for network certificates}
Putting this altogether, let $\delta > 0$, let $N>0$ be the number of estimates needed to calculate a certificate, and let $m$ be the number of geometric estimators to take a maximum over. Then with probability $(1-\delta)$, if we bound the tail probability for each geometric estimate with $\hat{\delta} = \left(\frac{\delta}{N}\right)^{1/m}$, then we have an upper bound on the certificate. 

\paragraph{MNIST example}
As an example, suppose we use the MNIST network from \cite{wong2017provable}. Then, let $\delta = 0.01$, $m=10$, and note that $N=6572$. Then, $\hat{\delta} = 0.26$, which we can achieve by using $k=200$ and $\epsilon=0.22$.

\begin{figure}[h]
\centering
\includegraphics[scale=0.4]{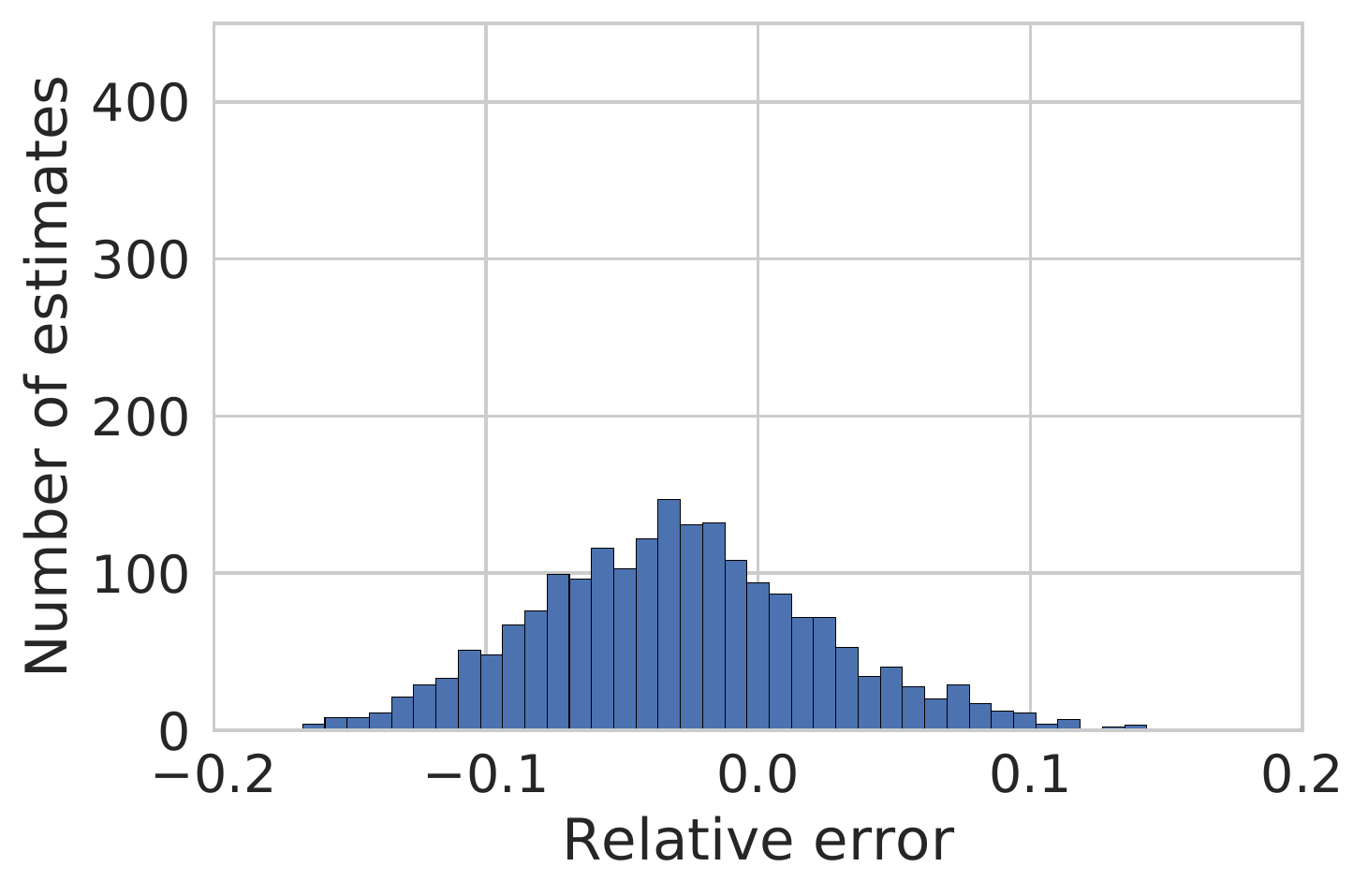}
\includegraphics[scale=0.4]{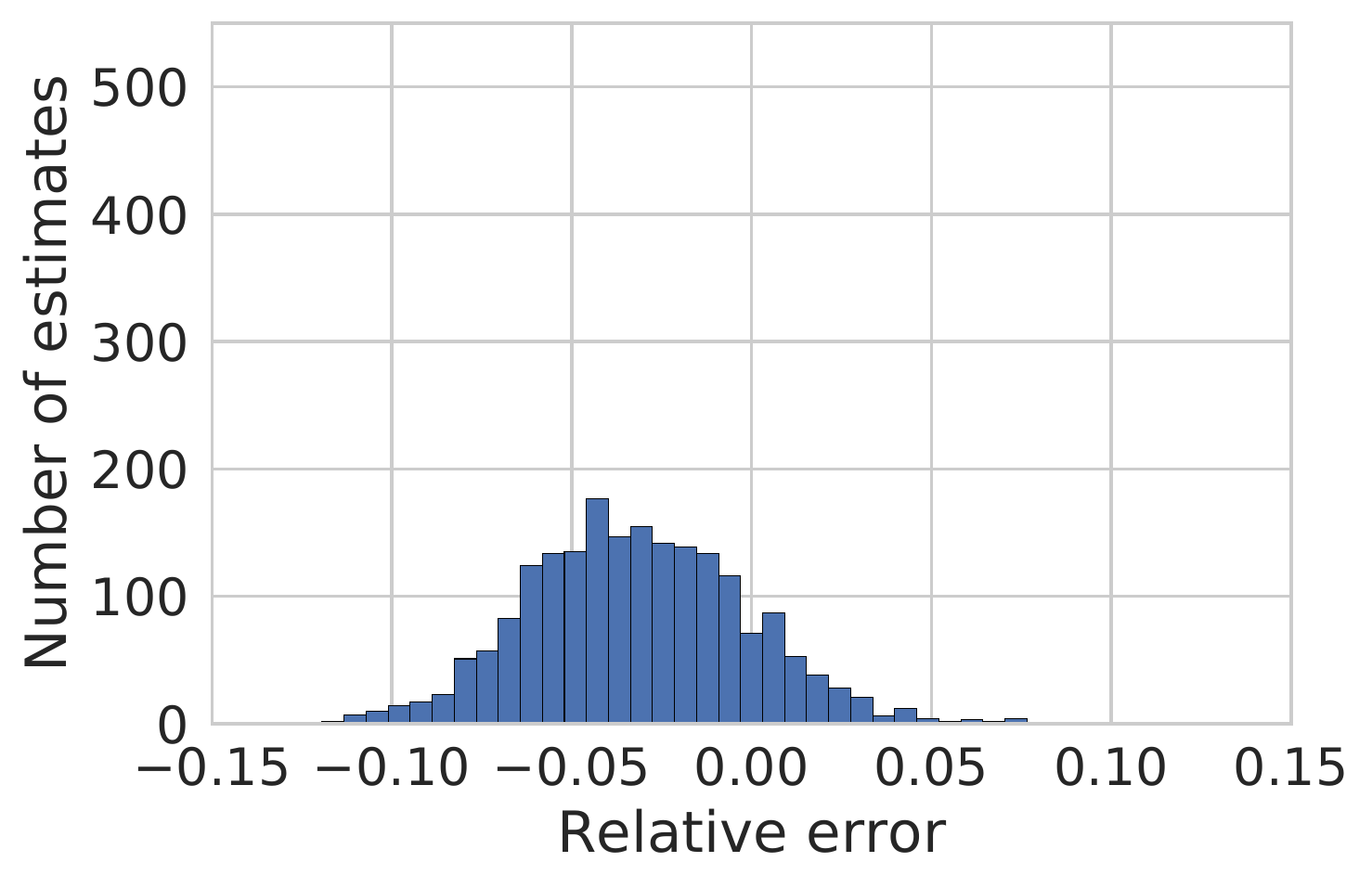}

\includegraphics[scale=0.4]{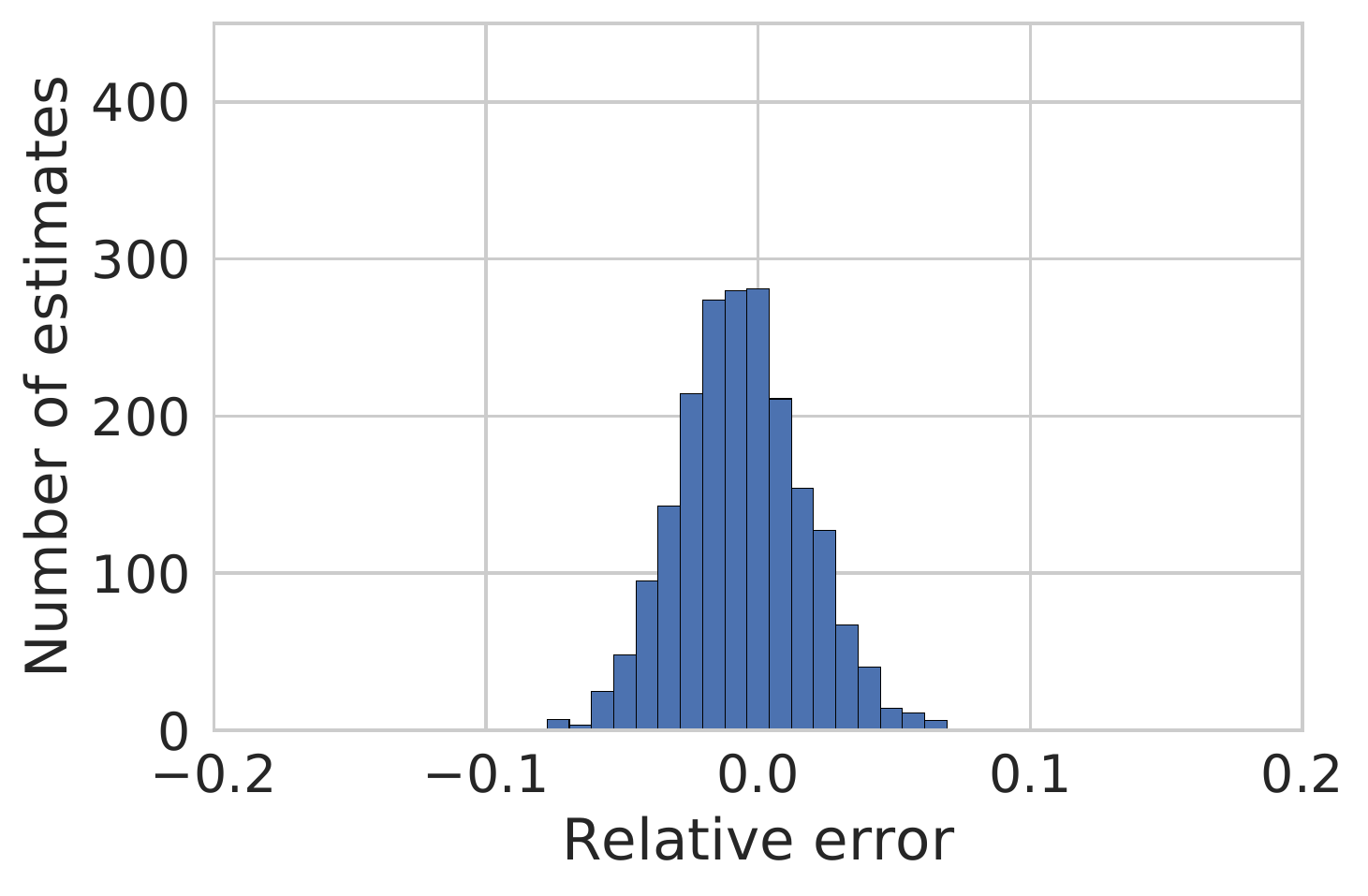}
\includegraphics[scale=0.4]{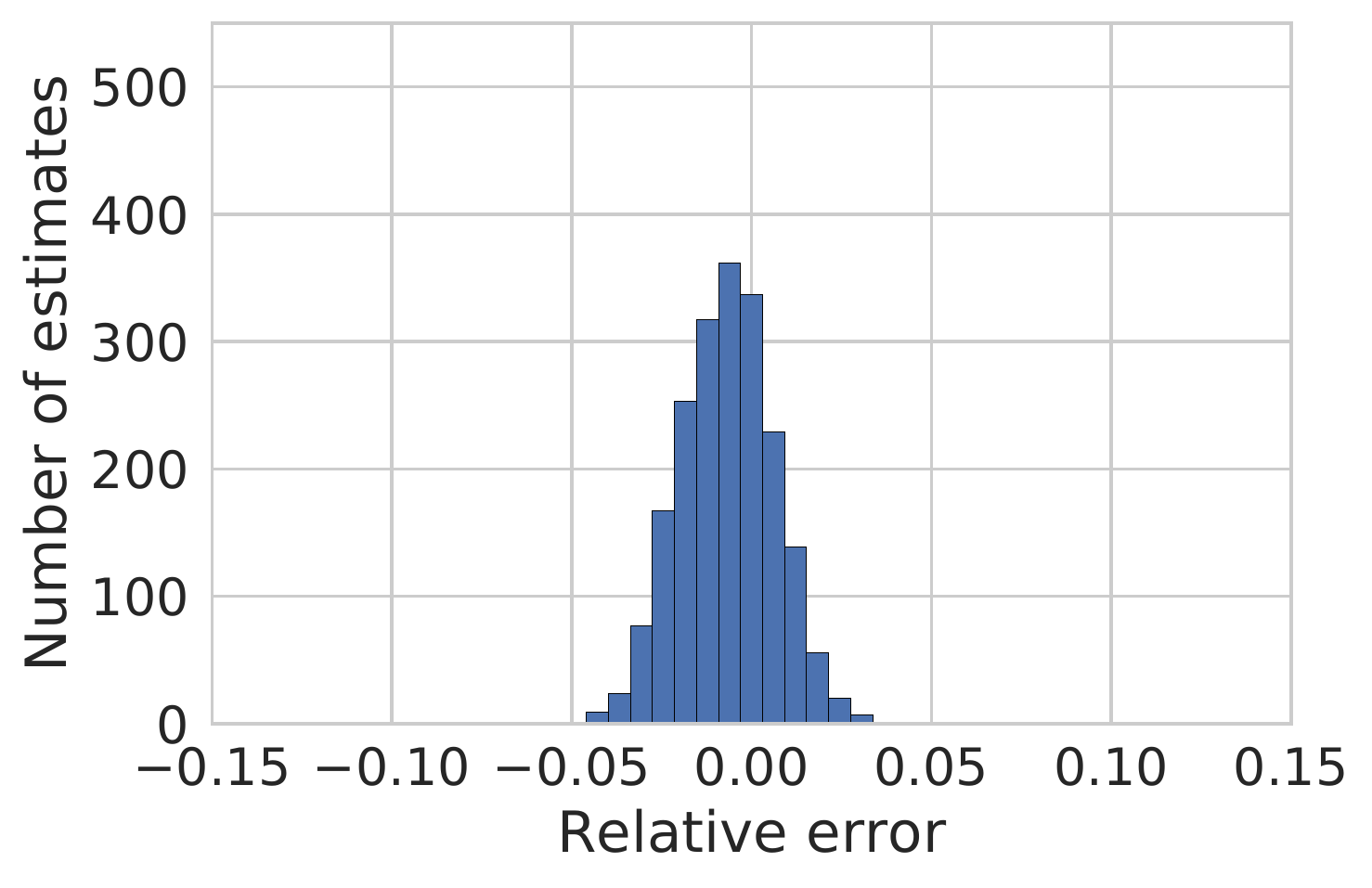}

\includegraphics[scale=0.4]{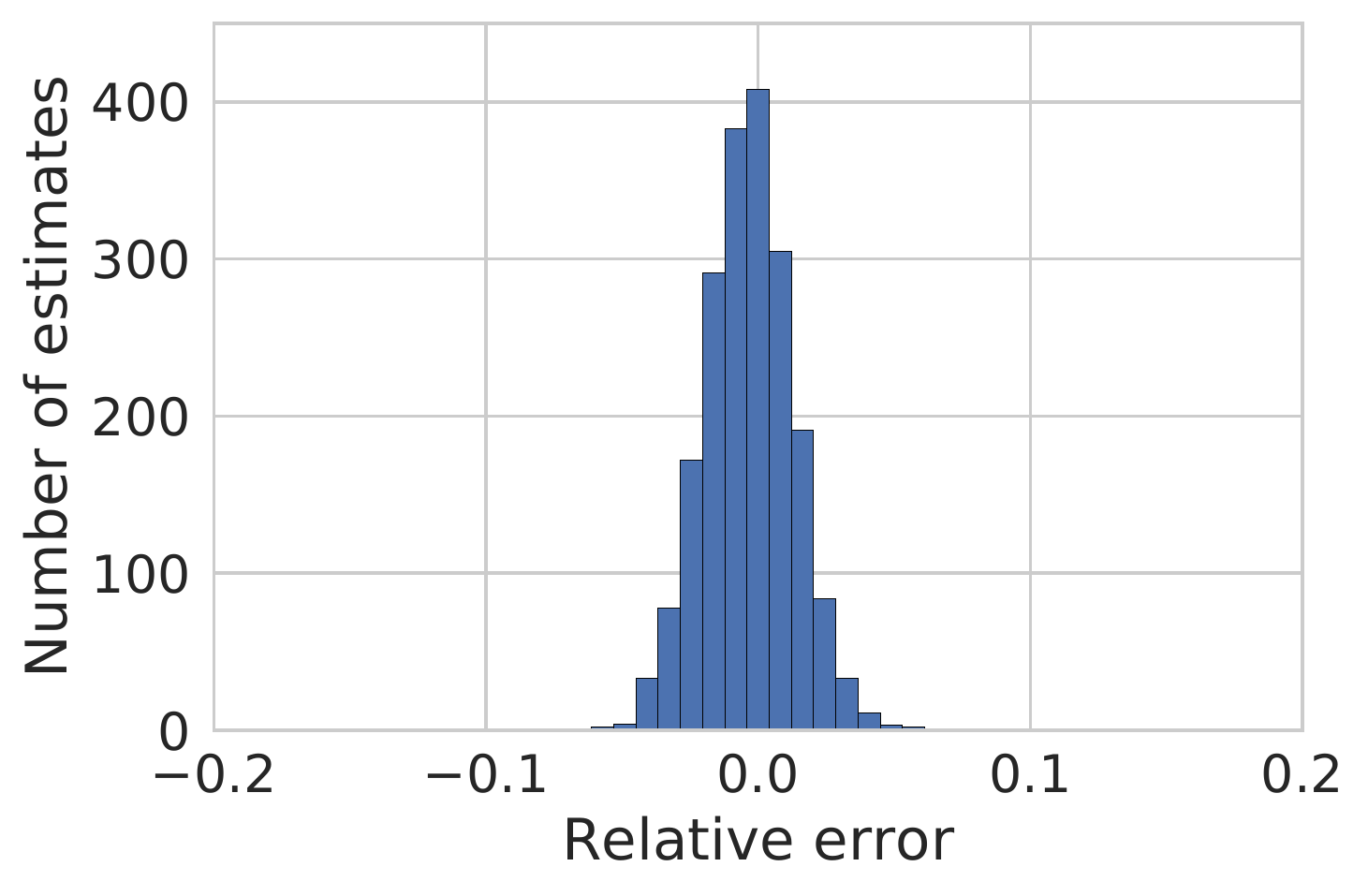}
\includegraphics[scale=0.4]{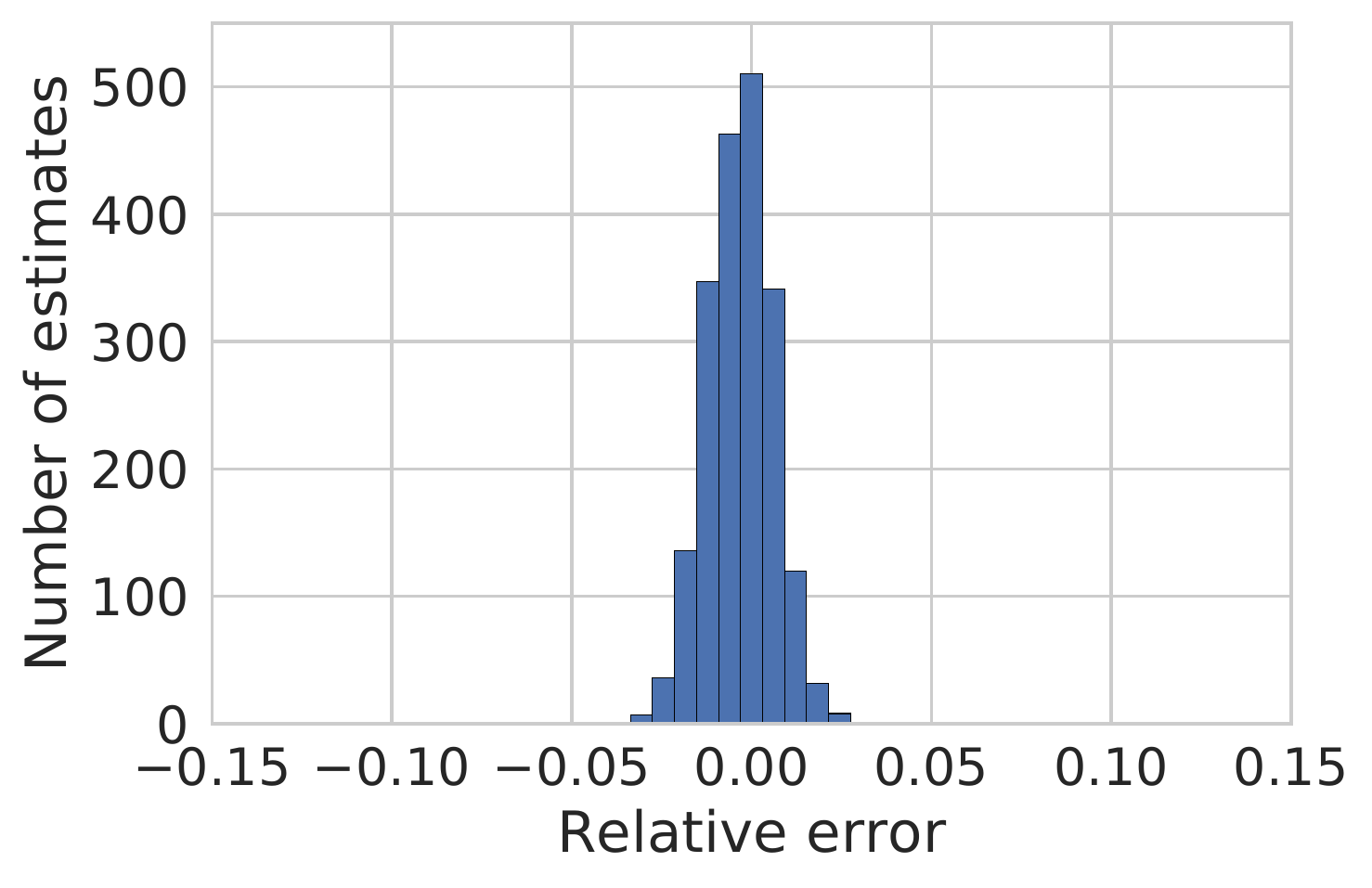}
\caption{Histograms of the relative error of the median estimator for 10 (top), 50 (middle), and 100 (bottom) projections, for a (left) random and (right) robustly trained convolutional layer.}
\label{fig:est}
\end{figure}

\subsection{Estimation quality and speedup}
\label{app:cauchy}
In this section, we discuss the empirical quality and speedup of the median 
estimator for $\ell_1$ estimation (for a more theoretical understanding, 
we direct the reader to \citet{li2007nonlinear}). In Figure \ref{fig:est}, we plot the relative 
error of the median estimator for varying dimensions on both an untrained 
and a trained convolutional layer, and see that regardless of whether the model is trained or not, 
the distribution of the estimate is normally distributed with decreasing variance for larger 
projections, and without degenerate cases. This matches the theoretical results derived in \citet{li2007nonlinear}. 

\begin{figure}
\centering
\includegraphics[scale=0.4]{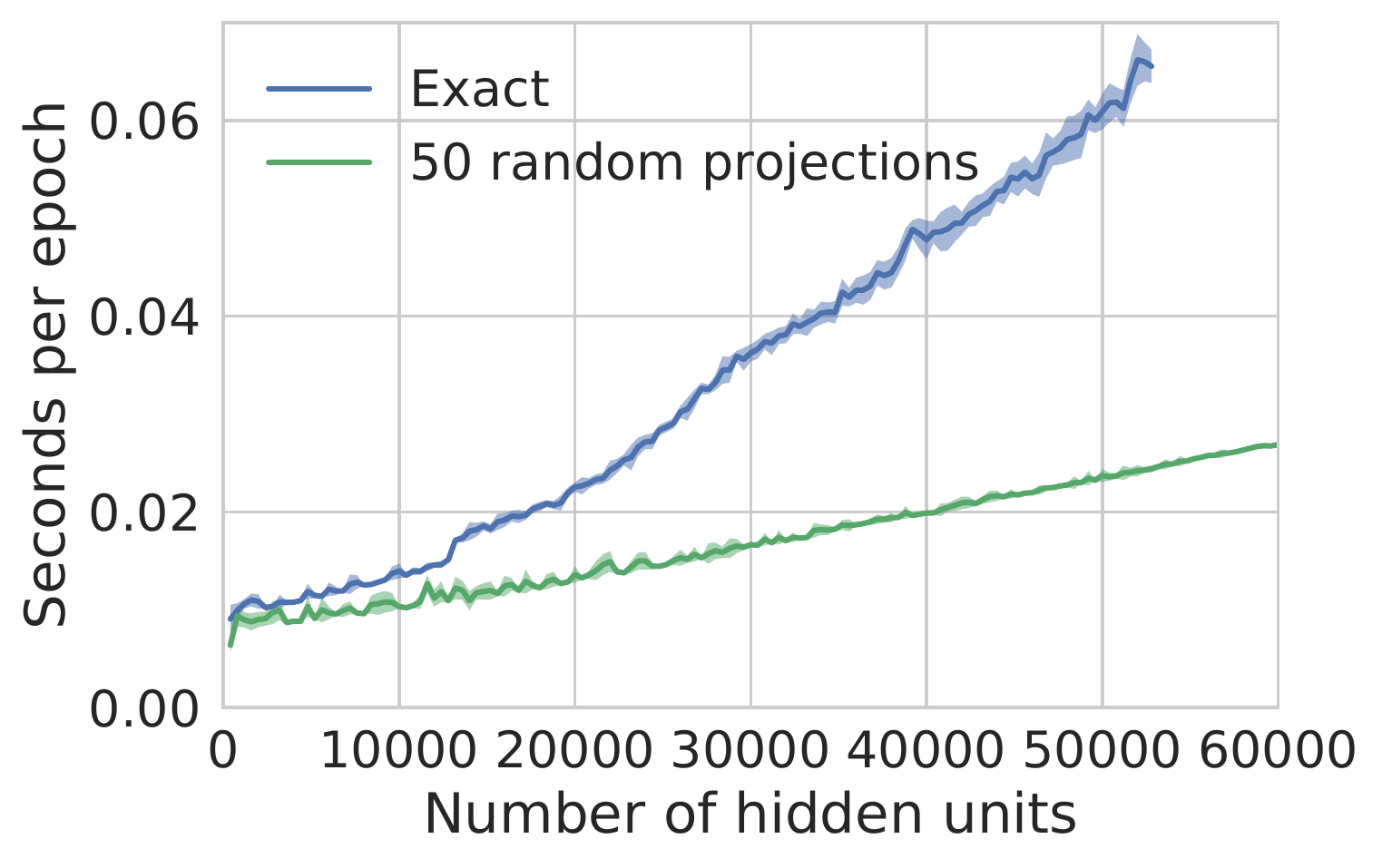}
\includegraphics[scale=0.4]{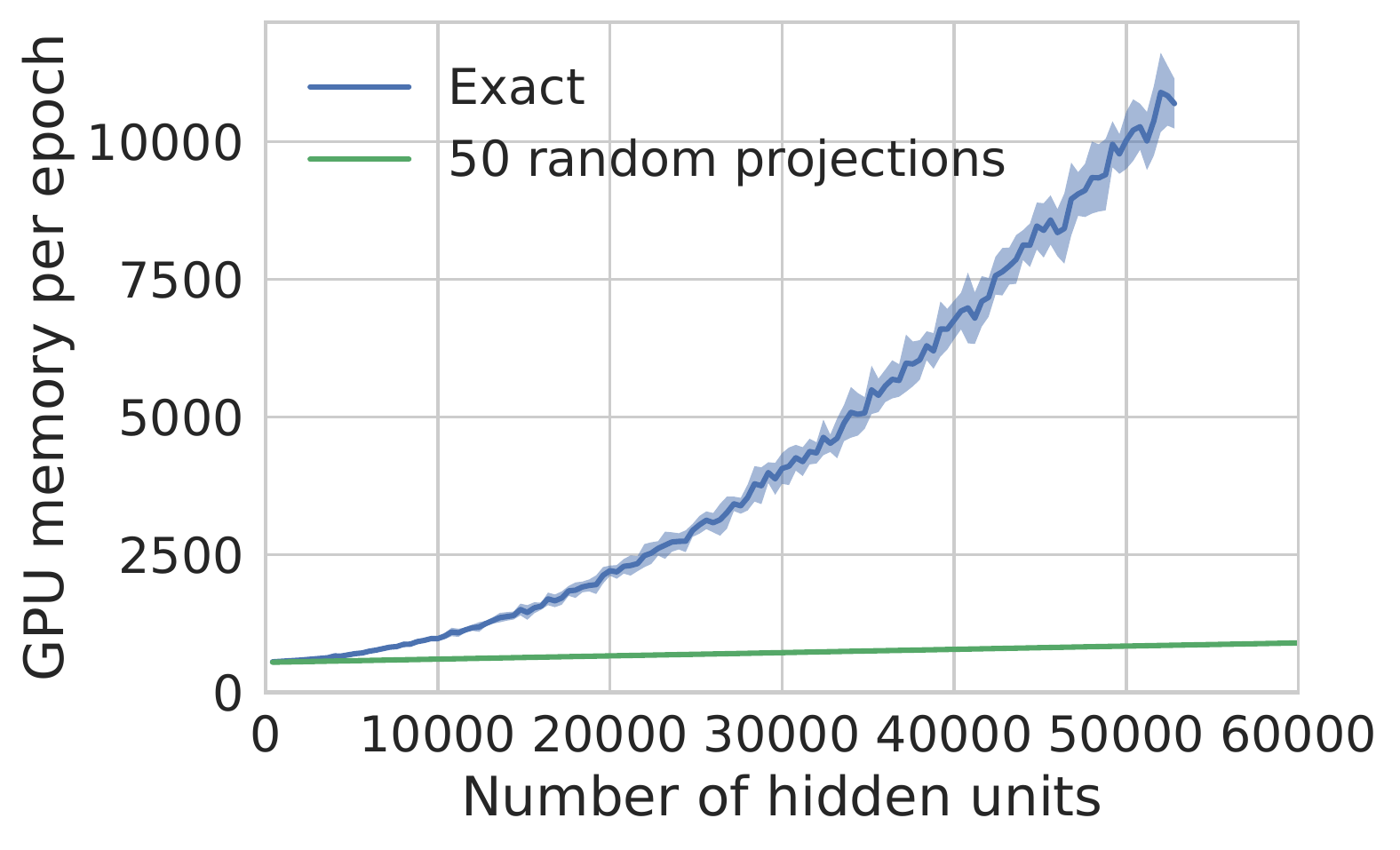}
\caption{Timing (top) and memory in MB (bottom) plots for a single 3 by 3 convolutional layer to evaluate 10 MNIST sized examples with minibatch size 1, averaged over 10 runs. The number of hidden units is varied by increasing the number of filters. On a single Titan X, the exact method runs out of memory at 52,800 hidden units, whereas the random projections scales linearly at a slope of $2.26\times 10^{-7}$ seconds per hidden unit, up to 0.96 seconds for 4,202,240 hidden units.}
\label{fig:metrics}
\end{figure}

In Figure \ref{fig:metrics}, we benchmark the time and memory usage on a 
convolutional MNIST example to demonstrate the performance improvements. 
While the exact bound takes time and memory that is quadratic in the number of
hidden units, the median estimator is instead linear, allowing it to scale up to 
millions of hidden units whereas the exact bound runs out of memory out at 50,280 hidden units. 

\section{AutoDual}
\label{app:autodual}
In this section, we describe our generalization of the bounds computation algorithm
from \citep{wong2017provable} to general networks using dual layers, which we call AutoDual. 

\paragraph{Efficient construction of the dual network via linear dual operators}
The conjugate form, and consequently the dual layer, for certain activations requires knowing lower and upper bounds for 
the pre-activations, as was done for ReLU activations in Algorithm 1 of 
\citet{wong2017provable}. 
While the bound in Equation \ref{eq:dual_obj} can be immediately used to compute all the bounds 
on intermediate nodes of the network one layer at a time, this requires performing a 
backwards pass through the dual network whenever we need to compute the bounds. 
However, if the operators $g_{ij}$ 
of the dual layers are all affine operators $g_{ij}(\nu_{i+1}) = A_{ij}^T\nu_{i+1}$ for some 
affine operator $A_{ij}$, we can apply a generalization of the lower and upper 
bound computation 
found in \citet{wong2017provable} to compute all lower and upper bounds, and 
consequently the dual layers, of the entire network 
with a single forward pass in a layer-by-layer fashion. With the lower and upper 
bounds, we can also use the same algorithm to automatically construct the 
dual network. The resulting algorithm, which we call AutoDual, is described in 
Algorithm \ref{alg:bounds}. 

In practice, we can perform several layer-specific enhancements on top of this algorithm. 
First, many of the 
$A_{ji}$ operators will not exist simply because most architectures (with a few exceptions) don't 
have a large number of skip connections, so these become no ops and can be ignored. Second, we can lazily 
skip the computation of layer-wise bounds until necessary, e.g. for constructing the dual layer 
of ReLU activations. Third, since many of the functions $h_j$ in the dual layers 
are functions of $B^T\nu_i$ for 
some matrix $B$ and some $i \geq j$, we can initialize 
$\nu_i^{(i)}$ with $B$ instead of the identity matrix, 
typically passing a much smaller matrix through 
the dual network (in many cases, $B$ is a single vector). 
%

\begin{algorithm}[tb]
   \caption{Autodual: computing the bounds and dual of a general network}
   \label{alg:bounds}
\begin{algorithmic}
  \STATE \textbf{input:} Network operations $f_{ij}$, data point
  $x$, ball size $\epsilon$
  \STATE \emph{// initialization}
  \STATE $\nu_{1}^{(1)} := I$
  \STATE $\ell_2 := x - \epsilon$ 
  \STATE $ u_2 := x + \epsilon$
  \FOR{$i=2,\ldots,k-1$}
  
  \STATE \emph{// initialize new dual layer}
  \STATE Create dual layer operators $A_{ji}$ and $h_i$ from $f_{ji}, \ell_j$ and $u_j$ for all $j\leq i$  
  \STATE $\nu_{i}^{(i)} := I$. 
  \STATE \emph{// update all dual variables}
  \FOR{$j=1, \dots, i-1$}
  \STATE $\nu_{j}^{(i)} := \sum_{k=1}^{j-1} A_{ki}\nu_{j}^{(k)}$
  \ENDFOR
  \STATE \emph{// compute new bounds}
  \STATE $\ell_{i+1} := x^T\nu_1^{(i)} - \epsilon\|\nu_1^{(i)}\|_{:} + \sum_{j=1}^{i}h_j(\nu_j^{(i)}, \dots,  \nu_{i}^{(i)}) $ 
  \STATE $ u_{i+1} := x^T\nu_1^{(i)} + \epsilon\|\nu_1^{(i)}\|_{:} - \sum_{j=1}^{i}h_j(-\nu_j^{(i)}, \dots,  -\nu_{i}^{(i)}) $
  \STATE \emph{// $\|\cdot\|_{:}$ for a matrix here denotes the norm of all rows}
  \ENDFOR
  \STATE \textbf{output:} bounds $\{\ell_i,u_i\}_{i=2}^{k}$, dual layer operators $A_{jk}, h_i$
\end{algorithmic}
\end{algorithm}

\section{Experiments} 

\label{app:results}
In this section, we provide more details on the experimental setup, as well
as more extensive experiments on the effect of model width and model depth on 
the performance that were not mentioned above. 

We use a parameter $k$ to control the width and depth of the architectures
used in the following experiments. 
The 
Wide($k$) networks have two convolutional layers of $4\times k$ and 
$8\times k$ filters followed by a $128\times k$ fully connected 
layer. The Deep($k$) networks have $k$ convolutional filters with 8 
filters followed by $k$ convolutional filters with 16 filters. 

\paragraph{Downsampling}
Similar to prior work, in all of our models 
we use strided convolutional layers with 4 by 4 kernels 
to downsample. When downsampling is not needed, we use 3 by 3 kernels 
without striding.

\subsection{MNIST}
\paragraph{Experimental setup}
For all MNIST experiments, we use the Adam optimizer with a learning rate of 0.001
with a batch size of 50. We schedule $\epsilon$ starting from 0.01 to the desired value over the first 20 epochs, after which we decay the learning rate by a factor of 0.5 every
10 epochs for a total of 60 epochs. 

\paragraph{Model width and depth}
We find that increasing the capacity of the model 
by simply making the network deeper and wider 
on MNIST is able boost performance. 
However, when the model becomes overly wide, the test
robust error performance begins to degrade due to overfitting. These results are shown in Table \ref{table:mnist}. 
\begin{table}
\centering
\caption{Results on different widths and depths for MNIST}
\label{table:mnist}

\tabcolsep=0.11cm
\begin{tabular}{lllllrr}
\textbf{Dataset} & \textbf{Model} & \textbf{Epsilon} & \textbf{Robust error} & \textbf{Error} \\
\hline
MNIST & Wide(1) & 0.1 & 6.51\% & 2.27\% \\
MNIST & Wide(2) & 0.1 & 5.46\% & 1.55\% \\
MNIST & Wide(4) & 0.1 & 4.94\% & 1.33\% \\
MNIST & Wide(8) & 0.1 & 4.79\% & 1.32\% \\
MNIST & Wide(16) & 0.1 & 5.27\% & 1.36\% \\
MNIST & Deep(1) & 0.1 & 5.28\% & 1.78\% \\
MNIST & Deep(2) & 0.1 & 4.37\% & 1.28\% \\
MNIST & Deep(3) & 0.1 & 4.20\% & 1.15\% \\
\hline
\end{tabular}
\end{table}

%
%

\subsection{CIFAR10}
\paragraph{Experimental setup} For all CIFAR10 experiments, we use 
the SGD optimizer with a learning rate of 0.05
with a batch size of 50. We schedule $\epsilon$ starting from 0.001 to the desired value over the first 20 epochs, after which we decay the learning rate by a factor of 0.5 every
10 epochs for a total of 60 epochs.

\section{Results for $\ell_2$ perturbations}
\label{app:l2}
We run similar experiments for $\ell_2$ perturbations on the input 
instead of $\ell_\infty$ perturbations, which amounts to replacing the $\ell_1$ 
norm in the objective with the $\ell_2$ norm. This can be equivalently scaled 
using random normal projections \citep{vempala2005random} instead of random Cauchy 
projections. We use the same network architectures as before, and 
pick $\epsilon_{2}$ such that the 
volume of an $\ell_{2}$ ball with radius $\epsilon_2$ is approximately the same as the volume of an $\ell_\infty$ ball with radius $\epsilon_\infty$. 
A simple conversion (an overapproximation within a constant factor) is: 
$$\epsilon_{2} = \sqrt{\frac{d}{\pi}}\epsilon_{\infty}.$$
For MNIST, we take an equivalent volume to $\epsilon_\infty=0.1$. This ends up being $\epsilon_2 = 1.58$, 
and note that within the dataset, the minimum $\ell_2$ distance between any two digits is at least 3.24, so $\epsilon_2$ is roughly half of the minimum distance between any two digits. For CIFAR we take an equivalent volume to 
$\epsilon_\infty=2/255$, which ends up being $\epsilon_2 = 36/255$. 

The results for the complete suite of experiments are in Table \ref{table:l2_results}, and we get similar
trends in robustness for larger and cascaded models to that of $\ell_\infty$ perturbations. 

\begin{table}
\centering
\caption{Results on MNIST, and CIFAR10 with small networks, large networks,
residual networks, and cascaded variants for $\ell_2$ perturbations.}
\label{table:l2_results}
\tabcolsep=0.11cm
\begin{tabular}{lll|rr|rr}
 &  &  & \multicolumn{2}{c|}{\textbf{Single model error}} & \multicolumn{2}{c}{\textbf{Cascade error}} \\
\textbf{Dataset} & \textbf{Model} & \textbf{Epsilon} & \textbf{Robust} & \textbf{Standard} &  \textbf{Robust} & \textbf{Standard}\\
\hline
MNIST & Small, Exact & 1.58 & 56.48\% & \textbf{11.86\%} & \textbf{24.42\%} & \textbf{19.57\%} \\
MNIST & Small & 1.58 & 56.32\% & 13.11\% & 25.34\% & 20.93\% \\
MNIST & Large & 1.58 & \textbf{55.47\%} & 11.88\%  & 26.16\% & 24.97\% \\
\hline
CIFAR10 & Small & 36/255 & 53.73\% & 44.72\%  & 50.13\% & 48.64\% \\
CIFAR10 & Large & 36/255 & 49.40\% & 40.24\%  & \textbf{41.36\%} & \textbf{41.16\%} \\
CIFAR10 & Resnet & 36/255 & \textbf{48.04\%} & \textbf{38.80\%}  & 41.44\% & 41.28\%\\
\end{tabular}
\label{fig:results}
\end{table}
\end{document}